\documentclass[letterpaper]{article} % DO NOT CHANGE THIS
\usepackage{aaai24}  % DO NOT CHANGE THIS
\usepackage{times}  % DO NOT CHANGE THIS
\usepackage{helvet}  % DO NOT CHANGE THIS
\usepackage{courier}  % DO NOT CHANGE THIS
\usepackage[hyphens]{url}  % DO NOT CHANGE THIS
\usepackage{graphicx} % DO NOT CHANGE THIS
\usepackage{natbib}  % DO NOT CHANGE THIS AND DO NOT ADD ANY OPTIONS TO IT
\usepackage{caption} % DO NOT CHANGE THIS AND DO NOT ADD ANY OPTIONS TO IT
\usepackage{algorithm}
\usepackage{algorithmic}

\usepackage{multirow}
\usepackage{caption}
\usepackage{subcaption}
\usepackage{ctable}
\usepackage{amsmath}
\usepackage{amssymb}

\newcommand{\etal}{\textit{et al}.}
\newcommand{\ie}{\textit{i}.\textit{i}.}

\usepackage{newfloat}
\usepackage{listings}
\DeclareCaptionStyle{ruled}{labelfont=normalfont,labelsep=colon,strut=off} % DO NOT CHANGE THIS
\floatstyle{ruled}
\newfloat{listing}{tb}{lst}{}
\floatname{listing}{Listing}
\nocopyright
\title{Boosting Diffusion Models with an Adaptive Momentum Sampler}
\author{Xiyu Wang, Anh-Dung Dinh, Daochang Liu, Chang Xu\\
    School of Computer Science, Faculty of Engineering,
    The University of Sydney, Australia\\
    {\tt\small xiyuwang.usyd@gmail.com, adin6536@uni.sydney.edu.au, } \\
    {\tt\small daochang.liu@sydney.edu.au, c.xu@sydney.edu.au}
}
\usepackage{bibentry}
\begin{document}

\maketitle

%%%%%%%%% ABSTRACT
\begin{abstract}
Diffusion probabilistic models (DPMs) have been shown to generate high-quality images without the need for delicate adversarial training. However, the current sampling process in DPMs is prone to violent shaking. In this paper, we present a novel reverse sampler for DPMs inspired by the widely-used Adam optimizer. Our proposed sampler can be readily applied to a pre-trained diffusion model, utilizing momentum mechanisms and adaptive updating to smooth the reverse sampling process and ensure stable generation, resulting in outputs of enhanced quality. By implicitly reusing update directions from early steps, our proposed sampler achieves a better balance between high-level semantics and low-level details. Additionally, this sampler is flexible and can be easily integrated into pre-trained DPMs regardless of the sampler used during training. Our experimental results on multiple benchmarks demonstrate that our proposed reverse sampler yields remarkable improvements over different baselines. We will make the source code available.
\end{abstract}

%%%%%%%%% BODY TEXT
\section{Introduction}
\label{sec:intro}
Deep generative modeling has emerged as a popular area of research due to its significance in comprehending and managing data. In image generation, GANs have dominated the field since its birth in \citeauthor{NIPS2014_5ca3e9b1}. Compared to other loglikelihood generative models \cite{kingma2013auto, hinton2006reducing, shao2021controlvae}, GANs show superiority over them in terms of both quality and diversity. However, the employment of adversarial training in GANs causes instability during its training and mode collapse, requiring specific optimization techniques and architectures. 

\begin{figure}
    \centering
    \begin{subfigure}[b]{0.23\textwidth}
        \centering
        \includegraphics[width=\textwidth]{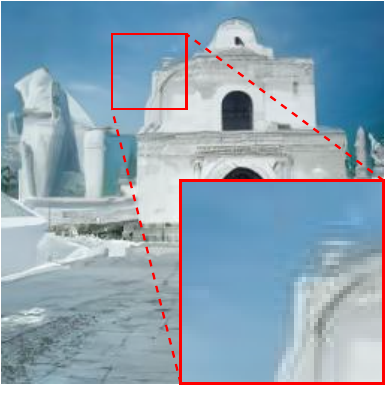}
        \caption{DDPM}
        \label{fig:example4noise-orign}
    \end{subfigure}
    \hfill
    \begin{subfigure}[b]{0.23\textwidth}
        \centering
        \includegraphics[width=\textwidth]{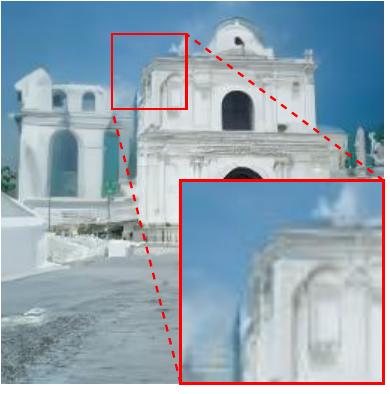}
        \caption{Our (DDPM + AM)}
        \label{fig:example4noise-our}
    \end{subfigure}
    \vskip -0.05in
    \caption{ Current DPMs' reverse sampling processes often exhibit severe vibration, which can lead to excessive noise or missing high-level components in the generated images. To address this issue, this paper proposes a new sampler with adaptive momentum (AM).}
    \vskip -0.2in
\end{figure}

As an alternative to adversarial training, iterative models such as Diffusion Probabilistic Models (DPMs) \cite{ho2020denoising, song2020score} have emerged as promising option. Denoising Diffusion Probabilistic Model (DDPM) \cite{ho2020denoising}, and its variants \cite{nichol2021improved, song2020denoising} are one of the streams based on the iterative process whose process includes two phases. The first phase is to diffuse the image into a predefined Gaussian noise, and the second phase will try to reverse the trajectory to recover the image. 
As a recent development, Denoising Diffusion Implicit Model (DDIM) \cite{song2020denoising} devises an accelerated version by turning DDPM into an implicit process and facilitates the reduction of the number of time steps during image generation. 
Meanwhile, the score-based scheme \cite{song2019generative, song2020improved, song2020score} is a theoretical version that works on the same mechanism. The main difference of the score-based model lies in its utilization of analytic tools, Stochastic Differential Equations (SDEs), to diffuse and denoise the image. 
The two iterative schemes of diffusion models and score-based models are unified upon the crucial connection observed in DDIM between
its optimization technique and Ordinary Differential Equations (ODEs). 
Therefore, we focus on the diffusion models in this paper since the two categories belong to the same broader generalization.

\begin{figure*}[t!]
     \centering
     \includegraphics[width=\textwidth]{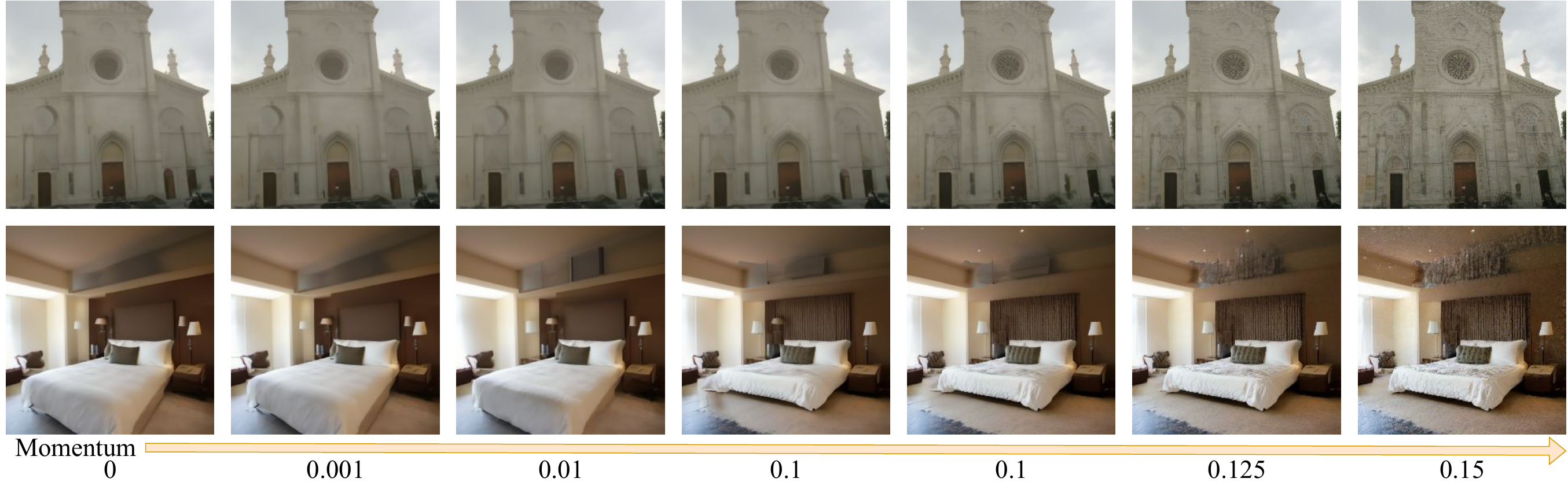}
     \caption{The images presented are generated from the final iteration, $\bar{x}_0$, after running 1000 steps of DDPM with our novel sampler with different scales of momentum. The momentum value ($b$) progressively increases from left to right (0 to 1.75), indicating an increasing reliance on earlier steps. With larger momentum, the final images prioritize intricate patterns and textures while also maintaining shapes and contours. By demonstrating the impact of varying momentum degrees, these final images provide an intuitive illustration of the adaptive momentum sampler's ability to improve low-level details while retaining high-level semantics.}
     \label{fig:example4momentum}
\vspace{-0.5cm}
\end{figure*}

The original reverse process in DPMs, based on the Markovian chain \cite{ho2020denoising}, can be regarded as one of the simple numerical solutions of the diffusion process’s Stochastic Differential Equations (SDEs). 
However, this vanilla reverse generative iteration used in DPMs, e.g., DDPM or DDIM, tends to meet narrow valleys, small humps, leading to excessive noise or missing high-level element in the generated images as the example in Fig.~\ref{fig:example4noise-orign}.
One reason for this is that the vanilla sampler used in DPMs employs noisy updates at each time step, which can lead to unstable sampling. Additionally, DDPM's use of a Markov process exacerbates this issue by deriving a new update direction at each step, without considering the global trajectory across steps. These limitations have motivated us to develop a new sampler that can more effectively explore the updating signals across steps, in order to stabilize the sampling process and facilitate the diffusion model's escape from local optima.
In this paper, we bring a new training-free reverse sampler for diffusion models to increase the quality of the generated samples by imitating the well-known Adam optimizer \cite{kingma2014adam}, which has shown excellent performance in neural network training.
Specifically, the proposed sampler acts in a non-Markovian manner by accumulating a velocity vector across steps to smooth the update direction.
A moving average of second-order moments is also maintained to adaptively adjust the update pace, which serves as a natural form of step-size annealing. 

\textit{The first benefit} of this new sampler is the foremost being a better trade-off between generating high-level semantics, such as shapes and outlines, and restoring low-level details, such as fine texture. 
To illustrate this point, we refer to Fig.~\ref{fig:example4momentum}, which depicts samples generated using different scales of momentum. 
As observed, employing a larger momentum that heavily re-uses the update direction of early steps leads to augmented details while still preserving high-level information. 
This phenomenon aligns with the two stages of learning for likelihood-based models \cite{rombach2022high}, where the first stage focuses on learning high-frequency details, while the second stage focuses on learning conceptual compositions. 
With a suitable momentum value, the adaptive momentum in our sampler coordinates these two learning stages, thus effectively balancing between high-level and low-level information.

\textit{The second benefit} of the proposed sampler is its flexibility to be easily integrated with existing pre-trained diffusion models. This non-parameterized sampler requires no additional training, and its momentum is dynamically adjusted on the fly during sampling. This adaptability is particularly advantageous in mitigating the train-test gap, as the sampler can be easily plugged into diffusion models pre-trained with different samplers or customized settings.

Experimental results on five common benchmarks,  CIFAR-10~\cite{krizhevsky2009learning}, CelebA~\cite{liu2018large}, ImageNet~\cite{deng2009imagenet}, LSUN~\cite{yu2015lsun} and CelebA-HQ~\cite{karras2017progressive}, show that the proposed adaptive momentum sampler improves both DDPM and DDIM in terms of generating quality.
In summary, the contribution of this paper is to newly propose a balanced, flexible, and highly-performing sampler that can be easily applied to pre-trained diffusion models.

\section{Related works}
\textbf{Diffusion Probabilistic Models} \cite{ho2020denoising} has been used as one of the effective generative models that avoid the adversarial training from GANs \cite{NIPS2014_5ca3e9b1}. By enabling the reverse process of diffusion, Denoising Diffusion Probabilistic Models (DDPMs) can reconstruct images. However, as previously mentioned, DDPMs suffer from the issue of expensive computation time due to their large iterative time steps. To address this problem, DDIM \cite{song2020denoising} "implicates" the model and allows it to run with far fewer iterations, significantly reducing the inference time compared to DDPM.

On the other hand, a score-based model via SDE is also a fresh approach to the diffusion model, where both the diffusion process and the denoising process are modeled by SDEs. Yang Song \etal \cite{song2019generative} first propose to generate samples from latent noise via the Dynamic Langevin Sampling technique. The author utilized the denoising score approximating scheme \cite{vincent2011connection} where the trajectory of noise is recorded in the scoring network $s_{\theta}$. The network is trained via Noise Conditional Score Networks. With the concurrent development of the DDPM/DDIM \cite{song2020denoising} along with the score-based models, and these two schemes both utilize diffusion and denoising schemes as their main mechanisms, Yang Song \cite{song2020score} proposes to model the diffusion process via a Stochastic Differential Equation (SDE) to unify the two streams. The diffusion process is modeled as the solution to an Ito SDE. The objective is to construct a diffusion process $\{ \mathbf{x}(t) \}^T_{t=0}$, the time index $t \in [0, T]$ is a continuous variable with $\mathbf{x}(0) \sim p_0$ belongs to the original data distribution.

\textbf{Improved Denoising Probabilistic Models}
\cite{dhariwal2021diffusion,dockhorn2021score,zheng2022entropy} propose to improve the quality of DDPM by using conditional information. While the achieved performance is significant, the two models need to train an additional classifier to achieve noise-aware information. Furthermore, the labels necessary for training these models are also exorbitant to collect in many practical tasks. In this work, we avoid using additional information besides training images. \cite{kim2022soft} improves the DDPM by softening the static truncation hyperparameters into a random variable. Ho \etal in \cite{ho2022cascaded} also propose cascaded diffusion models that show superior synthetic images compared to DDPM. However, these works require re-training the expensive DDPM to achieve little better results compared to the original baselines. In this work, we propose a momentum scheme that can behave on a pre-trained DDPM model. Thus, we could avoid large training DDPM/DDIM models, while still achieving fruitful results from improving image quality.

\section{Preliminary}
\label{sec:rw}

\subsection{Denoising Diffusion Probabilistic Model}

Generally, Gaussian diffusion models are utilized to approximate the data distribution $\mathbf{x}_0 \sim q(\mathbf{x}_0)$ by $p_{\theta}(\mathbf{x}_0)$. $p_{\theta}(\mathbf{x}_0)$ is modelled to be the form of latent variables models:
{
\small
\begin{align*}
\begin{split}
        p_{\theta}(\mathbf{x}_0) &= \int p_{\theta}(\mathbf{x}_{0:T})d\mathbf{x}_{1:T}, \text{where} \\
        p_{\theta}(\mathbf{x}_{0:T}) &= p_{\theta}(\mathbf{x}_T) \prod_{t=1}^{T} p_{\theta}^{(t)} (\mathbf{x}_{t-1}|\mathbf{x}_{t})
\end{split}
\end{align*}
}
where $x_1, x_2, ..., x_T$ are latent variables with the same dimensions with $x_0$.
{\small
\begin{alignat}{2}
    q(\mathbf{x}_{1:T} | \mathbf{x}_0) := & \prod_{t=1}^{T} q\left ( \mathbf{x}_t | \mathbf{x}_{t-1} \right ) \ , \text{where } \label{equ:q1} \\
    q\left ( \mathbf{x}_t \mid \mathbf{x}_{t-1} \right ) := & \mathcal{N} \left ( \mathbf{x}_t; \sqrt{1 - \beta_t}  \mathbf{x}_{t-1}, \beta_t \mathbf{I} \right ) \nonumber\ .
\end{alignat}}
Thus, the diffusion process from a data distribution to a Gaussian distribution for time step $t$ can be expressed as:
{\small
\begin{equation}\label{equ:samplext}
    \mathbf{x}_t = \sqrt{\alpha_t} \mathbf{x}_0 + \sqrt{1 - \alpha_t} \epsilon \ ,
\end{equation}}
where the $ \alpha_t := \prod_{i=0}^{t} \left ( 1 - \beta_i \right ) $ and $\epsilon \sim \mathcal{N} (\mathbf{0,I})$. The Ho \etal \cite{ho2020denoising} trains a U-net~\cite{ronneberger2015u} model $\theta$ to fit the data distribution $\mathbf{x}_0$ by maximizing the following variational lower-bound:
{\small
\begin{alignat}{2}
    & \max_\theta \mathbb{E}_{q\left(\mathbf{x}_0\right)}\left[\log p_\theta\left(\mathbf{x}_0\right)\right] \leq \label{equ:objective} \\
    & \max_\theta \mathbb{E}_{q\left(\mathbf{x}_0, \mathbf{x}_1, \ldots, \mathbf{x}_T\right)} \left[\log p_\theta\left(\mathbf{x}_{0: T}\right)-\log q\left(\mathbf{x}_{1: T} \mid \mathbf{x}_0\right)\right] \nonumber \ ,
\end{alignat}}
where the $q\left(\mathbf{x}_{1: T} \mid \mathbf{x}_0\right)$ is certain inference distribution, which could be calculated by the Bayes theorem with $\mathbf{x}_0$ and $\mathbf{x}_T$, over the latent variable.

\subsection{Denoising Diffusion Implicit Model}

DDPM’s dependence on the Markovian process has two main problems. First, due to the Markovian property, the generative process is forced to have a similar number of time steps to the diffusion process. Secondly, the stochastic process causes uncertainty in the synthetic images. This challenges the image interpolation in the latent space. 

DDIM generalizes the DDPM as a Non-Markovian process. Different from the Eq.~\ref{equ:q1}, $x_{t-1}$ is conditioned by both $x_t$ and $x_0$:
{\small
\begin{equation}\label{equ:ddimq}
    q_{\sigma}(\mathbf{x}_{1:T}|\mathbf{x}_0) = q_{\sigma}(\mathbf{x}_T|x_0) \prod_{t=2}^T q_{\sigma}(\mathbf{x}_{t-1}|\mathbf{x}_{t}, \mathbf{x}_0)
\end{equation}}

Where $q_{\sigma}(\mathbf{x}_{t-1}|\mathbf{x}_{t}, \mathbf{x}_0)$ is chosen so that $q_{\sigma}(x_T|x_0) = \mathcal{N}(\sqrt{\alpha_T} x_0, (1 - \alpha_T) \mathbf{I})$:
{\small
\begin{align*}
\begin{split}
        q_{\sigma}(\mathbf{x}_{t-1}|\mathbf{x}_t, \mathbf{x}_0) =& \mathcal{N}(\sqrt{\alpha_{t-1}}\mathbf{x_0} +\\
        &\sqrt{1 - \alpha_{t-1} - \sigma^2_t}\frac{\mathbf{x}_t - \sqrt{\alpha_t}\mathbf{x}_0}{\sqrt{1 - \alpha_T}}, \sigma_t^2 \mathbf{I})
\end{split}
\end{align*}}

From Eq.~\ref{equ:samplext}, $\mathbf{x}_t$ can be obtained by sampling $\mathbf{x}_0 \sim q(x_0)$ and $\epsilon_t \sim \mathcal{N}(0, \mathbf{I})$. By training the model $\epsilon_{\theta}^{(t)}$ to predict $\epsilon_t$ at each time step, $x_0$ predicting function $f_{\theta}^{(t)}$ is defined as:
{\small
\begin{equation*}
    f_{\theta}^{(t)}(\mathbf{x}_t)  = \frac{\mathbf{x}_t - \sqrt{1 - \alpha_t}\epsilon_{\theta}^{(t)} (\mathbf{x}_t)}{\sqrt{\alpha_T}} 
\end{equation*}}
This results in the whole generative process as:
{\small
\begin{equation*}
    p_{\theta}^{(t)} (\mathbf{x}_{t-1}|\mathbf{x}_{t}) = \begin{cases}
    \mathcal{N}(f_{\theta}^{(1)}(\mathbf{x}_1), \sigma_1^2\mathbf{I}) \quad & \text{if } t=1\\
    q_{\sigma}(\mathbf{x}_{t-1}| \mathbf{x}_{t}, f_{\theta}^{(t)}(\mathbf{x}_t)) & \text{otherwise},
    \end{cases}
\end{equation*}}
$\theta$ is optimized as Eq.~\ref{equ:objective} with $q_{\sigma}(\mathbf{x}_{1:T}|\mathbf{x}_0)$ is defined in Eq.~\ref{equ:ddimq}.
Based on the defined generative process, given a sample $\mathbf{x}_t$, we could sample the $\mathbf{x}_{t-1}$ as:
{\small
\begin{alignat}{2}
    \mathbf{x}_{t - 1} = & \sqrt{\alpha_{t - 1}} \underbrace{ \left ( \frac{\mathbf{x}_{t} - \sqrt{1 - \alpha_{t}} \epsilon_{\theta}^{(t)}(\mathbf{x}_{t})}{\alpha_{t}} \right )}_{\mathrm{predicted \ \mathbf{x}_0}} + \nonumber \\
    & \underbrace{\sqrt{1 - \alpha_{t - 1} - \sigma_t^2} \cdot \epsilon_{\theta}^{(t)}(\mathbf{x}_{t})}_{\mathrm{direction \ pointing \ to \ \mathbf{x}_t}} \ + \underbrace{\sigma_t \epsilon_t}_{\mathrm{random \ noise}}
    \label{eq:DDIM}
\end{alignat}}
where $\sigma_t = \eta \sqrt{(1 - \alpha_{t - 1})/(1 - \alpha_{t})} \sqrt{1 - \alpha_{t}/\alpha_{t - 1}}$ and $\eta=0$ \cite{song2020denoising} or $\eta=1$ \cite{song2020denoising} or $\eta = \sqrt{(1 - \alpha_{t})/(1 - \alpha_{t - 1})}$ \cite{ho2020denoising}. Yang Song \etal \cite{song2019generative} have proved that when $ \eta = \sqrt{(1 - \alpha_{t})/(1 - \alpha_{t - 1})} $ the Eq.~\ref{eq:DDIM} is essentially a different discretization to the same reverse-time SDEs and Jiaming Song \cite{song2020denoising} shows that when $ \eta = 0 $ the Eq.~\ref{eq:DDIM} similarity to Euler integration for solving ODEs. We provide when the $\eta = 1$ is the one type of DDPM, the reverse process is approximate solution of the same reverse-time SDEs in Appendix B.

\section{Methodology}
\label{sec:methodology}

In this paper, we propose a training-free Adaptive Momentum Sampler to generate high-quality images, which can build long-term communications between the previously explored path and the current denoising direction. The first part of this section presents how to integrate the basic momentum method into the origin inference process of DDPM/DDIM. After that, we illustrate how to effectively adjust the pacing rate for denoising with the moving average of the squared prior increments. In each part, we also show the design of how to trade-off between high- and low-level information.
\begin{figure}[t!]
     \centering
     \includegraphics[width=\linewidth]{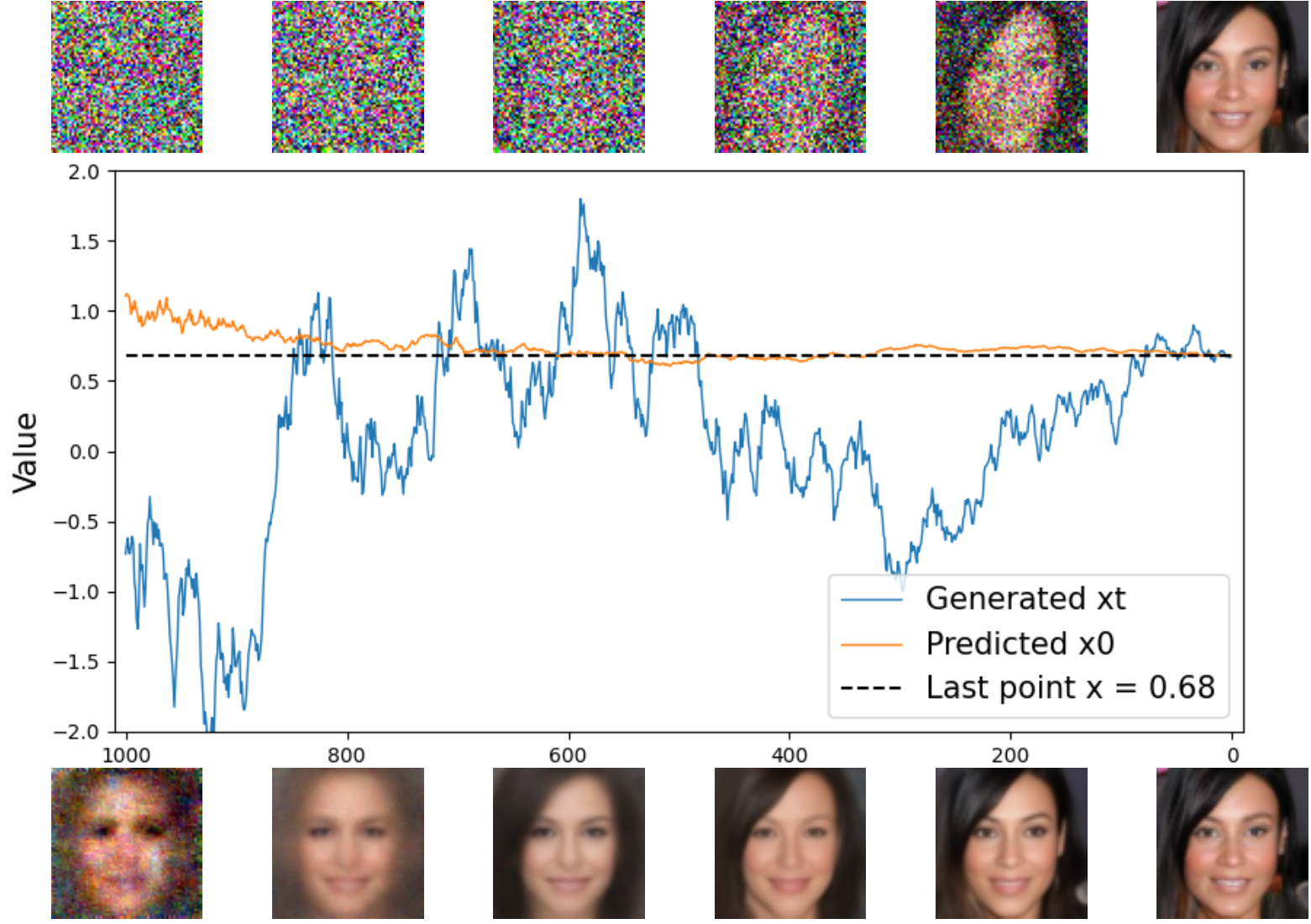}
     \vskip -0.1in
     \caption{ The line chart shows one pixel value of generated $\mathbf{x}_t$ (blue line) and the predicted $\mathbf{x}_0$ (orange line) in DDPM reverse process as Eq.~\ref{eq:DDIM}. The top and bottom pictures are the corresponding $t$-th generated $\mathbf{x}_t$ and predicted $\mathbf{\hat{x}}_0$.}
     \label{fig:example4methodology}
     \vskip -0.2in
\end{figure}
\subsection{Momentum Sampler}

We leverage the momentum method to achieve a better reverse process for diffusion models. The momentum method is widely used in accelerating the optimization of gradient flow in neural network training. 
The memorization of previously searched directions contributes to the optimizer's overcoming narrow valleys, small humps. 
We find that the momentum method can also benefit the reverse process of diffusion SDEs and ODEs. 

Inspired by SGD with momentum in optimizing neural networks, we replace the Euler-Maruyama method with the momentum method in the reverse process of diffusion models to avoid the plight mentioned. 
As shown in the Fig.~\ref{fig:example4methodology}, there is a phenomenon observed during the sampling phase: the process of reverse sampling $\mathbf{x}_t$ (blue line) is noisy with many vibrations, while the prediction of $\mathbf{x}_0$ from Eq.~\ref{eq:DDIM} (red line) is relatively stable and tractable. 
Therefore, the sampling process contains some trend information, enabling our momentum sampler to utilize such information and improve the sampling procedure 

However, the three terms in Eq.~\ref{eq:DDIM} are interdependent and complement each other, therefore we jointly track the history of these terms using a single momentum term. We first repeat the Eq.~\ref{eq:DDIM} as an incremental form of $\mathbf{x}$. For brevity, we let the $\bar{\mathbf{x}}_t = \ \frac{\mathbf{x}_{t}}{\sqrt{\alpha_{t}}}$ and the $\mu = \left( \sqrt{\frac{1 - \alpha_{t - 1} - \sigma_t^2}{\alpha_{t - 1}}} - \sqrt{\frac{1 - \alpha_{t}}{\alpha_{t}}} \right)$. And then, we can formulate the increment of $\bar{\mathbf{x}}_t$ and rewrite the DDPM (or DDIM) iteration Eq.~\ref{eq:DDIM}, as: 
\begin{alignat}{2}
\mathbf{d \bar{x}}_t =& \ \mu \cdot \epsilon_{\theta}^{(t)}(\mathbf{x}_{t}) + \frac{\sigma_t}{\sqrt{\alpha_{t - 1}}} \cdot \epsilon_t \ \mathrm{,}  \label{eq:d_x} \\
    \mathbf{\bar{x}}_{t-1} =& \ \mathbf{\bar{x}}_t + \mathbf{d \bar{x}}_t \ \mathrm{.} 
    \nonumber
\end{alignat}
Instead of only relying on the single last step result to generate samples, we incorporate the effects of prior directions as follows:
\begin{alignat*}{2}
    \mathbf{m}_{t-1} =& \enspace a \cdot \mathbf{m}_t + b \cdot \mathbf{d \bar{x}}_t 
    \ \mathrm{,}  \\
    \mathbf{\bar{x}}_{t-1} =& \enspace \mathbf{\bar{x}}_t + \mathbf{m}_{t-1} \ \mathrm{,} 
\end{alignat*}
where the $a$ is the damping coefficient, and the $a$ is a history-dependent velocity coefficient. Since the superposition property of the Gaussian distribution, we use the spherical difference, $ a^2 + b^2 = 1$, to balance the momentum and current value. The $\mathbf{m}_{t}$ is the current momentum of $t$-th iteration, and the $\mathbf{m}_{T} = \mathbf{0}$. The strength of the momentum method is controlled by the $a$ and $b$ coefficient.

\subsection{Adaptive Momentum Sampler}
In this section, we study how to adapt the learning rate based on the running average of recent magnitudes of the increment, $\mathbf{d \bar{x}}_t$. Besides the momentum-based sampling process only depending on the advantage of momentum by using a moving average, the idea of Root Mean Square Propagation \cite{mukkamala2017variants} (RMSProp) can also be applied to the probabilistic diffusion models generation process. To be more specific, we imitate RMSProp to automatically tailor the step size with a decaying average on each pixel. In this way, the process solver will focus on the most recently observed partial gradients and discard history from the extreme past as shown in Fig.~\ref{fig:example4methodology} blue line.

We propose an Adaptive Momentum sampling process that adapts the learning rates of each pixel per time step  The adaptive adjusting of denoising step size is based not only on the average first moment but also on the average of the second moments of the gradients. Based on the momentum sampling process, we use a $\mathbf{v_{t}}$ to store an exponentially decaying average of prior squared increment:
\begin{alignat}{3}
    \mathbf{v}_{t-1} \hspace{0.12cm} =& \enspace (1 - c) \cdot \mathbf{v_{t}} + c \cdot \left\| \mathbf{d \bar{x}}_t \right\|^2_2 \label{eq:adam_1} \\
    \mathbf{m}_{t-1} =& \enspace a \cdot \mathbf{m}_t + b \cdot \mathbf{d \bar{x}}_t \label{eq:adam_2} \\
    \mathbf{\bar{x}}_{t-1} =& \enspace \mathbf{\bar{x}}_t + \frac{\mathbf{m}_{t-1}}{\sqrt{\mathbf{v}_{t-1}} + \zeta } \ \mathrm{,} \nonumber
\end{alignat}
where the $c$ is the decay rate to control the moving averages of squared increment, and the $\mathbf{v_{t}}$ is the current averages of prior squared increment, and the $\mathbf{v_{T}} = \mathbf{1}$, and the $\zeta$ is a smoothing term (a small number) to avoid any division by 0. We use this moving average of second-order moments to make the reverse process concentrate on the current high-frequency information to avoid the image being over-smooth. Compared with the low-frequency information, the high-frequency information, such as the pattern and fine texture of images, is more sensitive to the second-order increment $\mathbf{d \bar {x}}$. In this assumption, the pacing rate of the high-frequency information will be adaptively magnified, while the pacing rate of the low-frequency information is relatively stable. The adaptive momentum sampler can future balance between high-frequency and low-frequency information.  

Algorithm \ref{alg:Adam} illustrates the adaptive momentum sampler process. Specifically, we use the momentum increment, $\mathbf{m}_{t-1}$, to replace the current increment of the $t$-th iteration of noised image. The $a_t$ and $b_t$ build a trade-off between the former and current time-dependent score function, demonstrating a strong numerical solver of the reverse process. As shown in Fig.~\ref{fig:example4momentum}, this adjustment can force the reverse sampling process to focus only on the early steps to generate more high-level semantics, such as shapes and outlines, and smooth the low-level information, such as the detailed pattern and texture. The $ c$ controls the exponential decay proportion for the second-moment estimates. In order to avoid excessive changes in the magnitude of the momentum increment for each pixel, we amass the square of the $ \mathbf{d \bar{x}}_t $'s $L_2$ norm and set the $\mathbf{v}_T=1$ during the implement. The basic momentum sampler could be considered as a particular case of the adaptive momentum solver when $c=1$ all the time. If $a=0$, $b=1$, and $c=1$  for all the $t$-th iterations, the sampling process would degenerate to the original DDPM or DDIM generation process.

\begin{figure}[!t]
\begin{center}
   \includegraphics[width=1.03\linewidth]{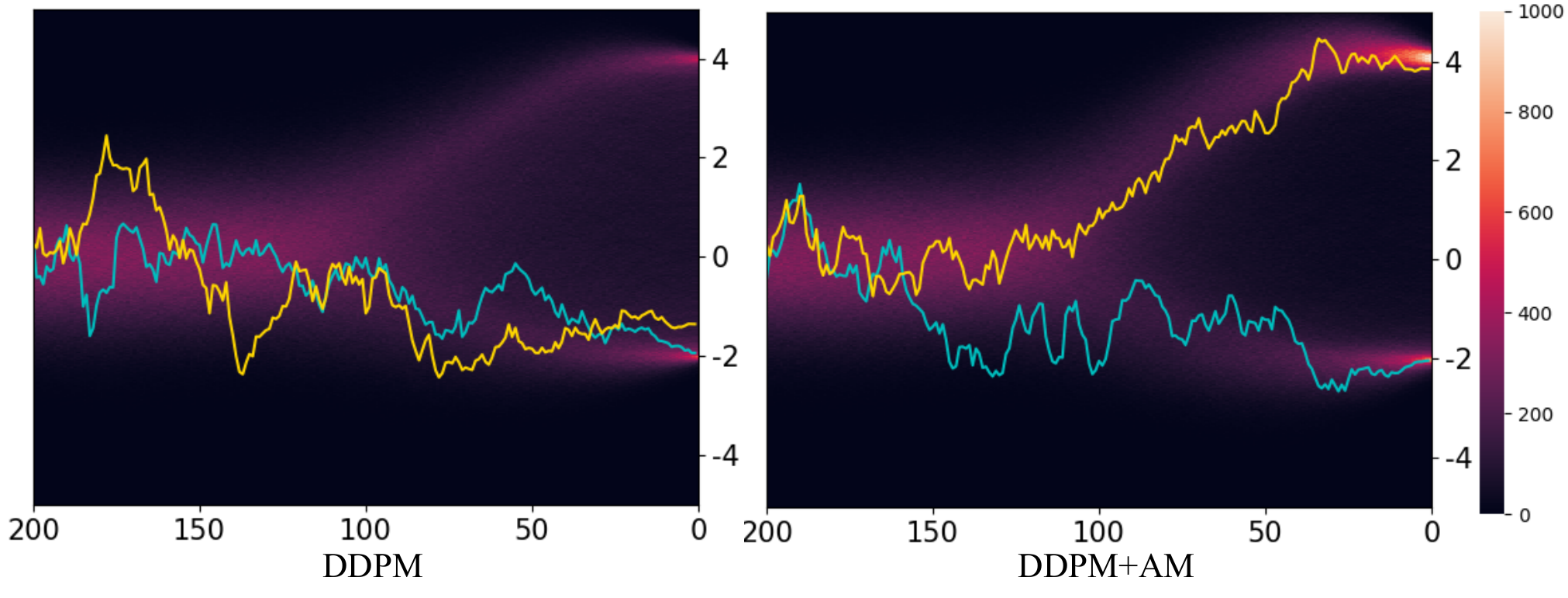}
\end{center}
\vskip -0.1in
\caption{This toy example on one-dimensional data to produce two points $x = -2$ and $x = 4$ in 200 steps. The background is the heat map with 10,000 samples. 
% The two lines is two  shows how the data value changes as the sampling process goes with the original DDPM sampler and our momentum sampler. 
Additionally, the cyan and gold lines represent the value of the generation sampling process with the original DDPM sampler and our momentum sampler.}
\label{fig:example4heatmap}
\vskip -0.25in
\end{figure}
\subsection{Synthetic Data and Analyze.} 
In this section, we will compare the generation process of the vanilla sampler with our proposed adaptive momentum sampler, both in practice and in theory.

\paragraph{Visualization} To perform a quantitative analysis, we trained a diffusion model to generate 1-dimension toy data with two points ($x=-2$ and $x=4$) over 200 timesteps. In Fig.~\ref{fig:example4heatmap}, we provide a visual representation of the differences between the original DDPM sampler and our proposed adaptive momentum sampler during the sampling process. The heatmaps in the background display the distribution of values for 10,000 samples generated by the original DDPM sampler and our adaptive momentum sampler, respectively. The lighter the color in the background, the greater the number of samples. 
{
% \vskip -0.1in
\begin{algorithm}[!tb]
    \renewcommand{\algorithmicrequire}{\textbf{Input:}}
    \renewcommand{\algorithmicensure}{\textbf{Output:}}
    \caption{Sampling with the Adaptive Momentum}    \label{alg:Adam}
    \begin{algorithmic}[1]
        \STATE{Initialization: $\mathbf{x_T} \sim \mathcal{N} (\mathbf{0,I}), \mathbf{m} = \mathbf{0}, \mathbf{v_{T}} = \mathbf{1}$}
        \FOR{$ t = T, \cdots , 1 $} 
            \STATE{$ \epsilon_t \sim \mathcal{N} \left ( \mathbf{0, I} \right ) $ if $t > 1$, else $\epsilon_t = \mathbf{0}$} 
            \STATE{Update $\mathbf{d \bar{x}}_t$ based on Equation~\ref{eq:d_x}}
            \STATE{$\mathbf{v}_{t-1} \hspace{0.12cm} = (1 - c) \cdot \mathbf{v_{t}} + c \cdot \left\| \mathbf{d \bar{x}}_t \right\|^2_2$}
            \STATE{$ \mathbf{m}_{t-1} = a \cdot \mathbf{m}_{t} + b \cdot \mathbf{d \bar{x}}_t$}
            \STATE{$\mathbf{x}_{t-1} \hspace{0.13cm}  = \sqrt{\alpha_{t - 1}} \left ( \frac{\mathbf{x}_{t}}{\sqrt{\alpha_{t}}} + \frac{\mathbf{m}_{t-1}}{\sqrt{\mathbf{v}_{t-1}} + \zeta }  \right )$}
        \ENDFOR
        \RETURN{$\mathbf{x}_{0}$}
    \end{algorithmic}
\end{algorithm}
% \vskip -0.1in
}
Specifically, the heat map generated by the adaptive momentum sampler is lighter, especially around the points $x=-2$ and $x=4$, indicating that it can more precisely produce the value of point. Furthermore, the generation trace of the adaptive momentum sampler is more stable, exhibiting less trembling, while the generation trace of the original DDPM sampler (gold line) features significant ineffective vibrations. Notably, for the gold line, the final value generated by the vanilla DDPM sampler is inaccurate. In contrast, the adaptive momentum sampler can more easily escape from narrow valleys, ensuring continuity in the generation process and producing a more accurate final value at last step.

\paragraph{Second-Order Approximation} We conducted a preliminary analysis to compare our method with the original method. 
As per Song et al.~\cite{song2020score}, the sampling process of DDPM/DDIM uses a first-order approximation of numerical SDEs/ODEs solvers. 
Building on this conclusion, we demonstrate in Appendix B that our method represents a second-order approximation of SDEs/ODEs. 
Instead of utilizing the Euler-Maruyama method for numerical solving SDEs/ODEs, we predict the next result by utilizing the midpoint method with previously generated results and the current state. 
This approach allows us to demonstrate that our method converges to the actual sample of the pre-trained model of DDPM.

In general, the adaptive momentum sampler typically employs historical data to facilitate a smoother generation process, as demonstrated by the gold line in Fig~\ref{fig:example4heatmap} for DDPM+AM. By contrast, the gold line in Fig~\ref{fig:example4heatmap} for DDPM illustrates early time steps characterized by wild swings, which can lead to misguided outcomes and large variance values for the final point.

\section{Experiments}
\label{sec:exper}

In this section, we quantitatively demonstrate the effectiveness of our proposed sampling method. First, we provide the experimental setup. Next, the method is quantitatively evaluated against the baseline samplers. The improvement compared to the baseline will be highlighted on different datasets. Lastly, an ablation study will be offered to inspect the impact of each hyper-parameter. 

\subsection{Experimental Setup}

\textbf{Datasets.} Following most of the setup in DDPM/DDIM and Latent Diffusion Models (LDM)~\cite{rombach2022high}, we utilize CIFAR10~\cite{krizhevsky2009learning} ($32\times32$), CelebA~\cite{liu2018large} ($64\times64$), ImageNet~\cite{deng2009imagenet} ($64\times64$), LSUN~\cite{yu2015lsun} ($256\times256$) and CelebA-HQ~\cite{karras2017progressive} ($256\times256$) in our experiments. 

\textbf{Configurations.} 
We mainly employ the original DDIM, DDPM, Analytic-DPM~\cite{bao2022analytic} and LDM in our comparisons as these schemes are solid and compact baselines. 
Nonetheless, our proposed schemes can be applied to other improved variants as well.  
For DDPM/DDIM and Analytic-DPM,  the same pre-trained diffusion models are used for the generation. The pre-trained models for CIFAR10, LSUN and CelebA-HQ are collected from the DDPM~\cite{ho2020denoising}, and the pre-trained models for CelebA and ImageNet are collected from DDIM~\cite{song2020denoising}, and IDDPM~\cite{song2020improved} respectively. 
Besides, we also use the pre-trained models from LDM~\cite{rombach2022high} for high-resolution image generation.
The sampling steps are set to 4000 on ImageNet and 1000 on other datasets. 
For the Eq.~\ref{eq:DDIM}, we select three $\eta$ values, the $\eta = 0$ will be equivalent to DDIM, $\eta=1$ will be a DDPM case in~\cite{song2020denoising}, and $\eta = \hat \eta$ is another DDPM case in its original paper \cite{ho2020denoising}. For settings where the sampling step does not equal the original pre-trained model, we follow the same strategy proposed in the DDIM, which replaces the $\alpha_{t}$ with the corresponding scaled $\alpha_{\tau}$. All the hyperparameters for the sampling process are presented in Appendix. Our experiments run on one node with 8 NVIDIA A100 GPUs.
Except for CelebA, the generation process is accelerated with the mixed-precision technique. 
We use `AM' to denote sampling with our proposed adaptive momentum scheme.

\begin{table}[!htpb]
    \vskip -0.1in
    \caption{The low-resolution image generation results on CIFAR10 (32 $\times$ 32), ImageNet (64 $\times$ 64) and CelebA (64 $\times$ 64) measured in IS $\uparrow$ and FID $\downarrow$.
    Note that a same pre-trained diffusion model is utilized on each dataset with different sampling strategies. 
    The improvement of the adaptive momentum sampling (`+AM') is remarkable on most of the datasets compared to the baselines in terms of FID.
    It is reasonable for our method to have similar IS to baselines, since the adaptive momentum mainly smooth the sampling trajectory which have no impact on changing the labels of the image, while IS is sensitive to inter-class diversity only.
    }
    \centering
    \vskip -0.1in
    \resizebox{\linewidth}{!}{
        \begin{tabular}{l|cccccc}
            \toprule[2pt]
            & \multicolumn{2}{c}{CIFAR10} & \multicolumn{2}{c}{ImageNet} & CelebA \\
            Sampler & IS & FID & IS & FID & FID \\
            \midrule[2pt]
            DDIM ($\eta = 0$) & 8.16 $\pm$ 0.15 & 3.98 & 16.40 $\pm$ 0.25 & 19.09 & 3.40 \\
            DDIM + AM & 8.16 $\pm$ 0.15 & \framebox[1.1\width]{3.81} & 16.35 $\pm$ 0.28 & \framebox[1.1\width]{18.85} & \framebox[1.1\width]{3.24} \ML[0.08em] 
            Analytic-DDIM  & 8.32 $\pm$ 0.09 & 4.66 & 16.29 $\pm$ 0.30 & 17.63 & 3.26 \\  
            Analytic-DDIM + AM & 8.30 $\pm$ 0.10 & \framebox[1.1\width]{\textbf{3.50}} & 16.35 $\pm$ 0.28 & \framebox[1.1\width]{\textbf{17.51}} & \framebox[1.1\width]{\textbf{3.14}} \\
            \midrule[2pt]
            DDPM ($\eta = 1$) & 8.22 $\pm$ 0.09 & 4.72 & 17.14 $\pm$ 0.20 & 16.55 & 5.78 \\
            DDPM + AM & 8.41 $\pm$ 0.09 & \framebox[1.1\width]{\textbf{3.53}} & 17.14 $\pm$ 0.21 & \framebox[1.1\width]{16.32} & \framebox[1.1\width]{2.50} \ML[0.08em]
            Analytic-DDPM & 8.51 $\pm$ 0.09 & 4.01 & 17.16 $\pm$ 0.21 & 16.42 & 5.21 \\
            Analytic-DDPM + AM & 8.52 $\pm$ 0.09 & \framebox[1.1\width]{\textbf{3.53}} & 17.15 $\pm$ 0.22 & \framebox[1.1\width]{\textbf{16.19}} & \framebox[1.1\width]{\textbf{2.29}} \\
            \midrule[2pt]
            DDPM ($\eta = \hat{\eta}$) & 8.39 $\pm$ 0.15 & 3.16 & 17.15 $\pm$ 0.18 & 16.38 & 3.26  \\
            DDPM + AM  & 8.46 $\pm$ 0.08 & \framebox[1.1\width]{\textbf{3.03}} & 17.15 $\pm$ 0.19 & \framebox[1.1\width]{\textbf{16.05}} & \framebox[1.1\width]{\textbf{3.17}} \\
            \bottomrule[2pt]
        \end{tabular}
}
\vskip -0.15in
\label{tab:low_res_exp}
\end{table}
\begin{table}[!htpb]
    % \vskip -0.05in
    \caption{The high-resolution image generation results on CelebA-HQ (256 $\times$ 256) , Church (256 $\times$ 256) and Bedroom (256 $\times$ 256) measured in FID $\downarrow$.
    The improvement of the adaptive momentum sampling (`+AM') is remarkable on all of the datasets compared to the baselines in terms of FID.
    }
    \centering
    \vskip -0.1in
    \resizebox{0.9\linewidth}{!}{
        \begin{tabular}{l|ccc}
            \toprule[2pt]
            Sampler & CelebA-HQ & Church & Bedroom \\
            \midrule[2pt]
            DDIM ($\eta = 0$) & 10.53 & 10.84 & 7.39 \\
            DDIM + AM & \framebox[1.1\width]{9.73} & \framebox[1.1\width]{8.17} & \framebox[1.1\width]{5.91} \ML[0.08em] 
            LDM-DDIM & 9.29 & 3.98 & 3.87 \\
            LDM-DDIM + AM & \framebox[1.1\width]{\textbf{9.13}} & \framebox[1.1\width]{\textbf{3.77}} & \framebox[1.1\width]{\textbf{3.54}} \\
            \midrule[2pt]
            DDPM ($\eta = 1$) & 12.34 & 7.81 & 6.24\\
            DDPM + AM & \framebox[1.1\width]{10.97} & \framebox[1.1\width]{7.74} & \framebox[1.1\width]{5.54} \ML[0.08em] 
            LDM-DDPM & 10.79 & 3.99 & 3.28\\
            LDM-DDPM + AM & \framebox[1.1\width]{\textbf{10.47}} & \framebox[1.1\width]{\textbf{3.92}} & \framebox[1.1\width]{\textbf{3.21}} \\
            \bottomrule[2pt]
        \end{tabular}
}
\vskip -0.2in
\label{tab:high_res_exp}
\end{table}

\textbf{Measurements.} Similar to many other generative models \cite{NIPS2014_5ca3e9b1,bao2022analytic,nichol2021improved,ho2020denoising}, we mainly adopt two evaluation metrics which are Frechet Inception Score (FID) and Inception Score (IS) \cite{lucic2018gans, borji2019pros}. IS is highly correlated with human-annotators \cite{salimans2016improved}. Nevertheless, this measure is often referred to as a method to measure inter-class diversity and is less sensitive to the diversity of the images inside one label \cite{lucic2018gans, borji2019pros}. In contrast, FID can detect intra-class mode collapsing as well \cite{lucic2018gans}. As a result, FID is considered a better measure for image generation tasks. A lower FID value indicates better performance, while a larger IS is better. 
For each experiment, we draw 50K samples for evaluation.

\begin{figure*}[!htpb]
    \vskip -0.1in
    \centering
    \includegraphics[width=\textwidth]{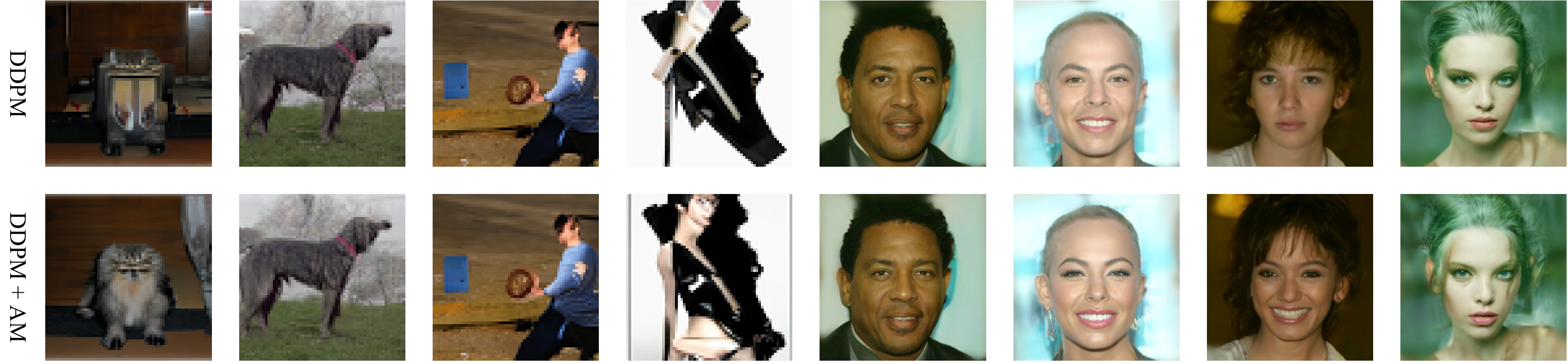}
    \vskip -0.05in
    \caption{Generated samples for ImageNet and CelebA-HQ}
    \label{fig:example4all}
\vskip -0.2in
\end{figure*}

\subsection{Overall Performance}

Our proposed sampling scheme is first compared with the baseline samplers of DDPM and DDIM in Table \ref{tab:low_res_exp} and \ref{tab:high_res_exp}. The same pre-trained diffusion DDPM/DDIM is used on each dataset to generate the data with vanilla  samplers and samplers combined with our adaptive momentum (the `+AM' rows). Analytic-DPM  \cite{bao2022analytic}, and LDM \cite{rombach2022high} sampler is also used as a baseline to justify our effectiveness on different diffusion schemes. 

% Low Res
The results in Table \ref{tab:low_res_exp} show the improvement of the adaptive momentum sampling over baselines on most low-resolution datasets (lower than 64x64) regarding FID.  For the CelebA dataset, the proposed method performs significantly better than DDPM with $\eta = 1$. It not only outperforms the baselines by around 50\% but also achieves the state-of-the-art for this dataset which reaches around 2.29 FID. For CIFAR10, although the FID has been very low for the baselines, we still have some rooms to improve from 3.16 to 3.03.

% High Res
Table \ref{tab:high_res_exp} shows that our adaptive momentum scheme leads to significant improvements in most baselines when applied to high-resolution datasets such as CelebA-HQ, LSUN Church, and LSUN Bedroom. We also demonstrate the compatibility of our approach with a popular high-resolution model, Latent Diffusion Models, which indicates that our method can be easily incorporated into other DPM-based methods. These results demonstrate the flexibility of our adaptive momentum scheme in improving the performance of various generative models.

Fig.~\ref{fig:example4all} presents a qualitative comparison of synthetic samples generated using different methods. In comparison to the original DDPM sampler, our proposed adaptive momentum sampler produces images with more realistic details and pronounced object outlines, confirming its ability to balance high-level and low-level information more effectively. For instance, our method enhances image details while preserving original patterns, such as the recovered person structure in row 2 column 4 and the enhanced facial details (forehead hair) in the last picture of the 2nd row.

\subsection{Ablation Study}

This subsection investigates different aspects of the settings, including different time steps and hyperparameters' effects.% and the momentum's adaptive properties.

\begin{table}[!tpb]
    % \vskip -0.1in
    \caption{This table displays the FID $\downarrow$ scores for generation results obtained using varying numbers of sampling steps. }
    \centering
    \vskip -0.1in
    \resizebox{0.8\linewidth}{!}{
        \begin{tabular}{l|cccc}
            \toprule[2pt]
            & \multicolumn{4}{c}{Sampling Steps} \\
            Sampler & 25 & 50 & 100 & 1000 \\
            \midrule[2pt]
            DDIM ($\eta = 0$) & 6.24  & 4.77 & \textbf{4.25} & 3.98  \\
            % FastDPM-DDIM & - & \textbf{2.86} & \textbf{3.20} & \textbf{5.05} \\
            Analytic-DDIM & 9.96 & 6.02 & 4.88 & 4.66 \ML[0.08em]
            DDIM + AM & \textbf{6.15} & \textbf{4.76} & 4.38 & \textbf{3.81} \\
            \midrule[2pt]
            DDPM ($\eta = 1$) & 14.44 & 8.23 & 5.82 & 4.72 \\
            Analytic-DDPM  & 8.50 & 5.50 & 4.45 & 4.31 \ML[0.08em]
            DDPM + AM & \textbf{7.19} & \textbf{4.10}  & \textbf{3.61}  & \textbf{3.53} \\
            \bottomrule[2pt]
        \end{tabular}
}
\vskip -0.3in
\label{tab:cifar10}
\end{table}

\paragraph{Different Sampling Steps.} 
Table~\ref{tab:cifar10} illustrates the performance of our method when fewer sample steps are used for acceleration on CIFAR10.
The proposed method could perform better or be comparable when using different sampling steps.
On most of the settings, we can provide up to 50\%  performance enhancements over the original DDPM sampler.
It is observed that the improvement is significant on the larger number of sample steps while less evident on the smaller ones.
This makes sense, possibly because using a small number of sampling steps is not always sufficient for accurate momentum estimation and noise suppression.

\begin{table}[!tpb]
    % \vskip -0.1in
    \caption{This table presents the results of a hyperparameter experiment measuring FID $\downarrow$ for our method. The best results among different values of $b$ and $c$ are marked in boxes, and the best results of each full column are shown in bold.}
    \centering
    \vskip -0.1in
    \resizebox{0.9\linewidth}{!}{
        \begin{tabular}{l|cccc}
            \toprule[2pt]
            & \multicolumn{4}{c}{Sampling Steps} \\
            Sampler & 25 & 50 & 100 & 1000 \\
            \midrule[2pt]
            DDPM ($\eta = 1$) & 14.44 & 8.23 & 5.82 & 4.72 \\
            DDPM + AM ($b = 0.05$) & 14.43 & 8.13 & 5.71 & 4.68 \\
            DDPM + AM ($b = 0.1$) & 10.13 & 6.52 & 5.88 & 4.09 \\
            DDPM + AM ($b = 0.15$)  & \framebox[1.1\width]{7.25} & \framebox[1.1\width]{4.41} & \framebox[1.1\width]{3.65} & \framebox[1.1\width]{3.54} \\
            DDPM + AM ($b = 0.2$) & 20.26 & 5.07 & 4.04 & 3.98 \\
            \midrule[2pt]
            DDPM + AM ($c = 0.001$) & 7.23 & 4.39 & 3.64 & 3.54  \\
            DDPM + AM ($c = 0.005$) & 7.22 & 4.35& 3.62 & \framebox[1.1\width]{\textbf{3.53}}  \\
            DDPM + AM ($c = 0.01$) & \framebox[1.1\width]{\textbf{7.19}} & 4.33 & \framebox[1.1\width]{\textbf{3.61}} & \framebox[1.1\width]{\textbf{3.53}} \\
            DDPM + AM ($c = 0.1$) & 7.61 & \framebox[1.1\width]{\textbf{4.10}} & 3.67 & 3.85 \\
            \bottomrule[2pt]
        \end{tabular}
}
\vskip -0.3in
\label{tab:hyperparameters}
\end{table}

\paragraph{Hyper-Parameters \& Ablation Study.} 
Considering about the effects of $b$ and $c$ in the Eq.~\ref{eq:adam_1} and \ref{eq:adam_2}, we setup with different values as in Table~\ref{tab:hyperparameters} on CIFAR10. The first five lines illustrate how the FID of the generated images changes as the $b$ increases when $c=0$. In general, the image quality will first increase and then decrease. When the $b = 0.15$, the FID achieves the lowest value for image generation in different sampling steps, which is the best trade-off point. The last four lines show how the $b$ influence the sampling results when $b = 0.15$. Similarly, the FID first decreases and then increases when the $b$ decrease. The last four rows reflect the impact of $c$ with $b = 0.15$, where the most of best performance is obtained at $c=0.01$. Empirically, different denoising steps can share the same hyper-parameters as shown in Table 3 and Appendix. Thus, we only need to find $b_{\max}$ like {\small $10^{-3}, 10^{-2}$ or $ 10^{-1}$} in the small denoising steps, e.g., 100 or 200. Most of hyperparameter values are shared across different datasets. We suggest choosing {\small $b = 10^{-3}$ or $10^{-2}$ and $c=0.01$} for high-resolution images as default. 

Table~\ref{tab:hyperparameters} also presents the results of an ablation study on our proposed method, the Adaptive Momentum Sampler. The first five rows display the outcomes when the sampling process only utilizes the moving average of the moment increment.  Moreover, the comparison between the first and fourth rows reveals that our proposed method outperforms the original reverse process sampler, underscoring the efficacy of the momentum idea. Furthermore, comparing the fourth and eighth rows, we observe that introducing adaptivity through the moving average of second-order moments further enhances the performance of our adaptive momentum sampler, which demonstrates its excellent generalization ability. 

\section{Conclusion}

In this work, we propose a training-free sampler to improve the generated images of Diffusion Generative Models, especially for discrete discrete time steps settings. Through extensive experiments, we found out that the adaptive momentum sampler helps to smooth the denoising trajectory and, hence, improves the quality of the generated images. In future works, we will expand the scheme to continuous settings as well as with solid theoretical versions.

\bibliography{refer.bib}

\clearpage

\twocolumn[
\section*{\center{\LARGE{Appendix for ``Boosting Diffusion Models with an Adaptive Momentum Sampler''}}}
]

\vspace{0.1in}
In this section, we provide additional experiments, proof and generated samples about the adaptive momentum sampler.

\appendix

\section{Experiments}

In this section we show how we choose the hyperparameter and additional experiments on CIFAR10, ImageNet, CelebA, CelebA-HQ, LSUN bedroom and Church with different methods and the momentum's adaptive properties.

\subsection{Hyperparameter Details}

\begin{table}[!htpb]
    %\vskip -0.1in
    \caption{Hyperparameters setting for CIFAR10, ImageNet, CelebA with different methods}
    \centering
    %\vspace{0.2cm}
    \resizebox{\linewidth}{!}{
        \begin{tabular}{l|l|ccc}
            \toprule[2pt]
            Method & & CIFAR10 & ImageNet & CelebA \\
            \midrule[2pt]
            \multirow{2}{*}{
            \begin{tabular}[c]{@{}l@{}}
            DDIM + AM \\ ($\eta = 0$)
            \end{tabular}} & $b$ & 0.4 & $1 \times 10^{-3}$ & $1 \times 10^{-3}$ \\
            \cline{2-5}
            & $c$ & 0.9999 & 0.9999 & 0.9999 \\
            \cline{1-5}
            \multirow{2}{*}{
            \begin{tabular}[c]{@{}l@{}}
            Analytic-DDIM + AM 
            \end{tabular}} & $b$ & 0.4 & $1 \times 10^{-3}$ & $1 \times 10^{-3}$ \\
            \cline{2-5}
            & $c$ & 0.9999 & 0.9999 & 0.9999 \\
            \midrule[2pt]
            \multirow{2}{*}{
            \begin{tabular}[c]{@{}l@{}}
            DDPM + AM \\ ($\eta = 1$)
            \end{tabular}} & $b$ & 0.15 & 0.15 & 0.15 \\
            \cline{2-5}
            & $c$ & 0.999 & 0.999 & 0.9 \\
            \cline{1-5}
            \multirow{2}{*}{
            \begin{tabular}[c]{@{}l@{}}
            Analytic-DDPM + AM 
            \end{tabular}} & $b$ & 0.15 & 0.15 & 0.15 \\
            \cline{2-5}
            & $c$ & 0.999 & 0.999 & 0.9 \\
            \midrule[2pt]
            \multirow{2}{*}{
            \begin{tabular}[c]{@{}l@{}}
            DDPM + AM \\ ($\eta = \hat{\eta}$)
            \end{tabular}} & $b$ & 0.125 & $1 \times 10^{-3}$ & $1 \times 10^{-3}$ \\
            \cline{2-5}
            & $c$ & 0.999 & 0.999 & 0.9 \\
            \bottomrule[2pt]
        \end{tabular}
}
%\vskip 0.1in
\label{tab:hyper-low-res}
\end{table}
\begin{table}[!htpb]
    %\vskip -0.1in
    \caption{Hyperparameter setting for CelebA-HQ, LSUN Church and Bedroom with different methods.}
    \centering
    %\vspace{0.2cm}
    \resizebox{\linewidth}{!}{
        \begin{tabular}{l|l|cccc}
            \toprule[2pt]
            Method & & CelebA-HQ & Church & Bedroom \\
            \midrule[2pt]
            \multirow{2}{*}{
            \begin{tabular}[c]{@{}l@{}}
            DDIM + AM \\ ($\eta = 0$)
            \end{tabular}} & $b$ & $1 \times 10^{-3}$ & $3 \times 10^{-3}$  & $1.5 \times 10^{-3}$ \\
            \cline{2-5}
            & $c$ & 0.9999 & 0.9999 & 0.9999 \\
            \cline{1-5}
            \multirow{2}{*}{
            \begin{tabular}[c]{@{}l@{}}
            LDM-DDIM + AM 
            \end{tabular}} & $b$ & $1 \times 10^{-3}$ & $1 \times 10^{-3}$  & $1 \times 10^{-3}$ \\
            \cline{2-5}
            & $c$ & 0.999 & 0.999 & 0.999 \\
            \midrule[2pt]
            \multirow{2}{*}{
            \begin{tabular}[c]{@{}l@{}}
            DDPM + AM \\ ($\eta = 1$)
            \end{tabular}} & $b$ & $1 \times 10^{-2}$ & $1 \times 10^{-2}$ & $5 \times 10^{-2}$ \\
            \cline{2-5}
            & $c$ & 0.999 & 0.999 & 0.999 \\
            \cline{1-5}
            \multirow{2}{*}{
            \begin{tabular}[c]{@{}l@{}}
            LDM-DDPM + AM 
            \end{tabular}} & $b$ & $1 \times 10^{-3}$ & $1.5 \times 10^{-3}$  & $1 \times 10^{-3}$ \\
            \cline{2-5}
            & $c$ & 0.999 & 0.999 & 0.999 \\
            \bottomrule[2pt]
        \end{tabular}
}
%\vskip 0.1in
\label{tab:hyper-high-res}
\end{table}

Tables~\ref{tab:hyper-low-res} and~\ref{tab:hyper-high-res} provide an overview of the hyperparameters $b$ and $c$ used in our method to obtain the results presented in Tables~\ref{tab:low_res_exp} and~\ref{tab:high_res_exp}, respectively. Notably, we observ that the hyperparameter choices for the same dataset using a different method (DDPM/DDIM) are similar. Similarly, for the same pre-trained model and similar sampling method, such as "DDPM + AM" and "Analytic-DDPM + AM", the hyperparameter choices can be also similar.

Overall, we recommend using {\small $b_{\max} = 10^{-3}$ or $10^{-2}$ and $c=0.999$} as the default hyperparameters for achieving satisfactory improvements, with the exception of the CIFAR10 dataset.

\subsection{Different Combinations of $a$ and $b$}

\begin{figure*}[t!]
     \centering
     \includegraphics[width=\textwidth]{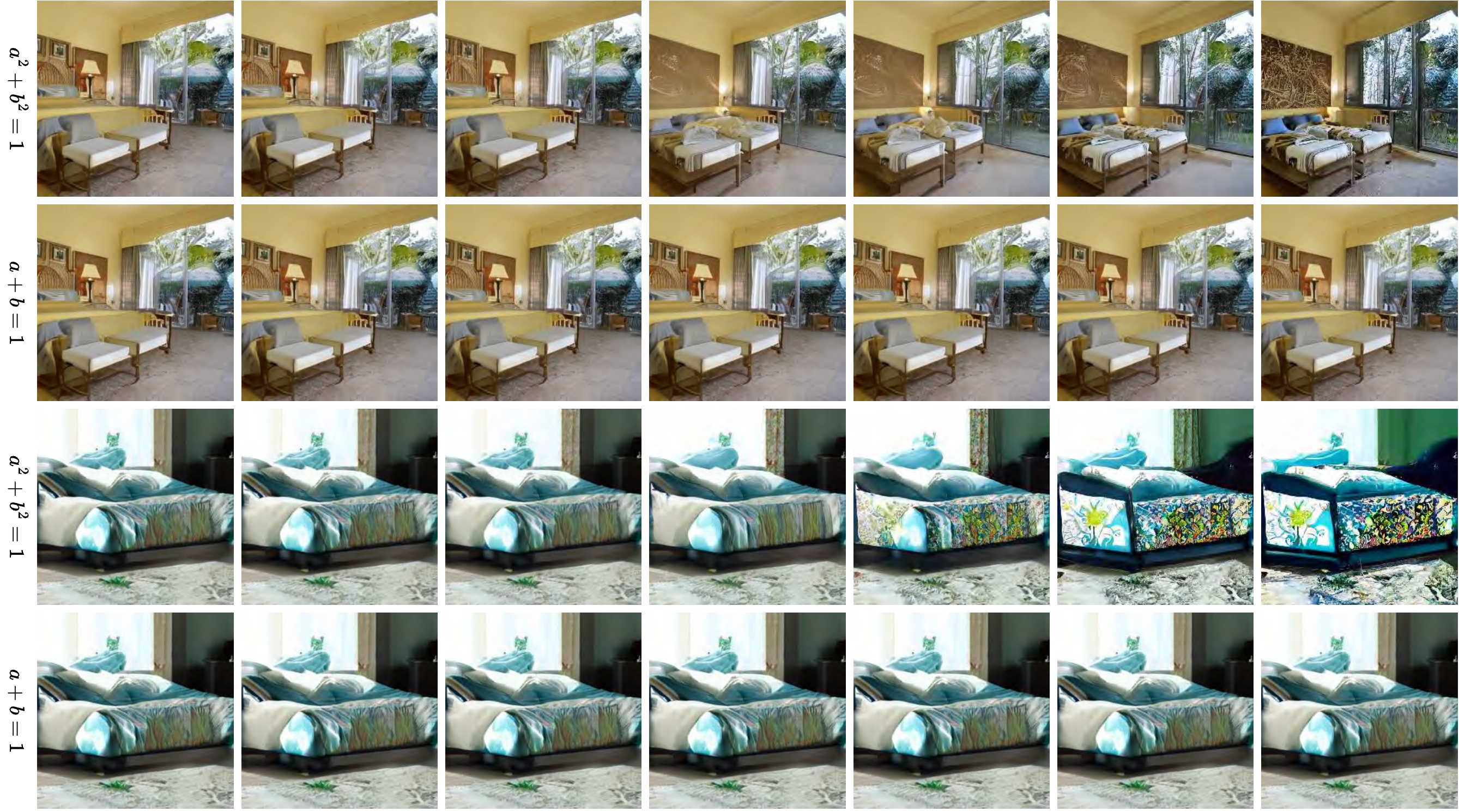}
     \caption{ Samples generated by DDPM-AM with different $a$ and $b$ combinations. From left to right, the $b_{\max}$ is increase from 0.005 to 0.2 for four line pictures}
     \label{fig:example4a&b}
\end{figure*}

In this subsection, we employ the generated samples to further demonstrate the reason behind our choice of $a^2 + b^2 = 1$, as illustrated in Fig.\ref{fig:example4a&b}. Specifically, we consider two combinations of $a$ and $b$: $a + b = 1$ or $a^2 + b^2 = 1$. At a high level, the momentum of $\bar{\mathbf{x}}$ can adjust the current search direction based on the previously explored path. When $a + b = 1$, as depicted in Fig.\ref{fig:example4a&b}, this adjustment can force the reverse sampling process to focus only on the early steps, emphasizing high-level semantics such as shapes and outlines while smoothing out low-level information such as patterns and textures. On the other hand, when $a^2 + b^2 = 1$, the sum of $a$ and $b$ exceeds 1 since $a, b < 1$. In this scenario, the momentum method can preserve or even enhance more detailed information, including low-level semantics as shown in Fig.~\ref{fig:example4a&b}. As a result, we typically choose $a^2 + b^2 = 1$ to strike a better trade-off between high-level and low-level semantics.
 
\subsection{Additional Different Sampling Steps Experiments}

Table~\ref{tab:bedroom} and Table~\ref{tab:church} show the performance of our method when fewer sample steps are used for acceleration.
The proposed method could achieve better or comparable performance when using different sampling steps.
As mentioned above, it is observed that the improvement is significant on the larger number of sample steps, while less evident on the smaller ones.
This makes sense possibly because using a small number of sampling steps is not always sufficient for accurate momentum estimation and noise suppression.

\begin{table}[!htbp]
    %\vskip -0.1in
    \caption{LSUN Bedroom image generation results measured in FID $\downarrow$, with different numbers of sumpling steps used.}
    \centering
    %\vspace{0.2cm}
    \resizebox{0.75\linewidth}{!}{
        \begin{tabular}{l|cccc}
            \toprule[2pt]
            & \multicolumn{4}{c}{Bedroom} \\
            dim($\tau$) & 25 & 50 & 100 & 1000 \\
            \midrule[2pt]
            DDIM ($\eta = 0$) & \textbf{7.75} & 6.54 & 6.52 & 7.39 \\
            FastDPM-DDIM & 9.86 & 8.37 & 9.94 & - \ML[0.08em]
            DDIM + AM  & 7.78 & \textbf{6.42} & \textbf{6.02} & \textbf{5.90} \\
            \midrule[2pt]
            DDPM ($\eta = 1$) & 19.12 & 10.96 & 6.91 & 6.24  \\
            FastDPM-DDPM & 20.12 & 10.12 & 7.98 & - \ML[0.08em]
            DDPM + AM  & \textbf{18.14} & \textbf{7.90} & \textbf{5.43} & \textbf{5.46} \\
            \bottomrule[2pt]
        \end{tabular}
}
%\vskip 0.1in
\label{tab:bedroom}
\end{table}
\begin{table}[!htbp]
    %\vskip -0.1in
    \caption{LSUN Church image generation results measured in FID $\downarrow$, with different numbers of sumpling steps used.}
    \centering
    %\vspace{0.2cm}
    \resizebox{0.75\linewidth}{!}{
        \begin{tabular}{l|cccc}
            \toprule[2pt] 
            & \multicolumn{4}{c}{Church} \\
            dim($\tau$) & 25 & 50 & 100 & 1000 \\
            \midrule[2pt]
            DDIM ($\eta = 0$) & 11.26 & 10.57 & 10.64 & 10.84 \ML[0.08em]
            DDIM + AM  & \textbf{10.62} & \textbf{9.17} & \textbf{8.67} & \textbf{8.17} \\
            \midrule[2pt]
            DDPM ($\eta = 1$) & \textbf{17.92} & 9.98 & 7.84 & 7.81 \ML[0.08em]
            DDPM + AM  & 17.95 & \textbf{9.11} & \textbf{7.57} & \textbf{7.74} \\
            \bottomrule[2pt]
        \end{tabular}
}
%\vskip 0.1in
\label{tab:church}
\end{table}

\subsection{Rate-Distortion Trade-Off}

\begin{figure}[!tpb]
\begin{center}
   \includegraphics[width=0.7\linewidth]{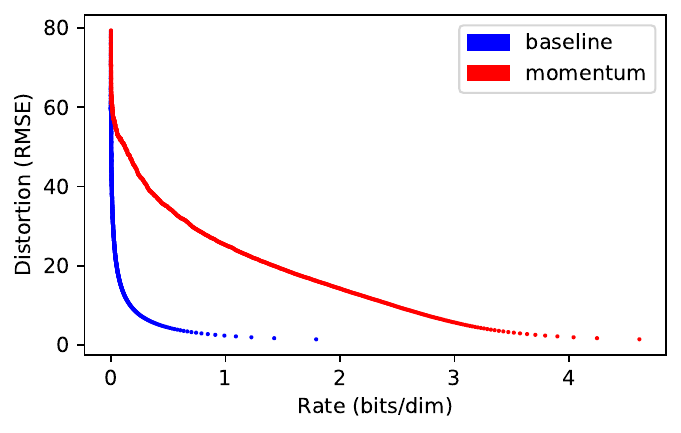}
\end{center}
% \vspace{-.5cm}
\vskip -0.1in
\caption{The rate-distortion trade-off on CIFAR-10. The proposed momentum scheme could achieve a better balance between high-level semantics and low-level details.}
\label{fig:rate-distortion}
\vskip -0.2in
\end{figure}

The rate-distortion curves of a trained model using both adaptive momentum inference and baseline inference schemes are plotted in Fig.~\ref{fig:rate-distortion}. 
It was criticized in the literature that diffusion models tend to spend much more rates, \ie, model capacity, on restoring imperceptible distortions than high-level semantics~\cite{ho2020denoising,rombach2022high}, which is also observed in Fig.~\ref{fig:rate-distortion} for the baseline scheme.
In contrast, the newly adaptive proposed momentum sampling is able to straighten the curve and thus strike a better balance between generating high-level semantics and low-level details.
This could provide a possible explanation for the improved generation performance observed in our experiments.
Note that it is reasonable for the red curve to have larger rates, which are computed by the cumulative sum of the variational bound terms~\cite{ho2020denoising}, since the momentum inference is being applied to a model pre-trained without momentum. 
We expect that incorporating an adaptive momentum strategy into the training process could further enhance the rate-distortion trade-off as well as the generation quality in future works.

\section{Proofs}

In this section, we provide proofs that the reverse process is a first-order approximation of the reverse-time SDEs, while our proposed method, the adaptive momentum sampler, constitutes a second-order approximation of the reverse-time SDEs/ODEs.

\subsection{Reverse Diffusion Sampling}

Our proposed method leverage the adaptive momentum to enhance the reverse solver. Thus, it is important to prove that the DDPM/DDIM sampler is a solution of the reverse-time SDEs/ODEs. Yang Song \etal \cite{song2019generative} have proved that when $ \eta = \sqrt{(1 - \alpha_{t})/(1 - \alpha_{t - 1})} $ the Eq.~\ref{eq:DDIM} is essentially a different discretization to the same reverse-time SDEs and Jiaming Song \cite{song2020denoising} shows that when $ \eta = 0 $ the Eq.~\ref{eq:DDIM} similarity to Euler integration for solving ODEs. 

Here, we provide when the $\eta = 1$ is the one type of DDPM, the reverse process (Eq.~\ref{eq:DDIM}) is approximate solution of the same reverse-time SDEs. According to the DDPM~\cite{ho2020denoising}, the forward process can be rewrite as:
\begin{equation}
    \mathbf{\bar{x}}_t = \mathbf{\bar{x}}_{t-1} + \sqrt{ \frac{\beta_t}{\alpha_t} } \ \epsilon \ \text{,}
\end{equation}
where the $\mathbf{\bar{x}}_t = \mathbf{x}_t / \alpha_t$, $\beta_t = 1 - \alpha_t / \alpha_{t-1}$ and $ \epsilon \sim \mathcal{N} (\mathbf{0,I}) $. Thus, based on~\cite{song2019generative}, the reverse-time SDE will be:
\begin{equation}
   \mathbf{\bar{x}}_{t-1} = \mathbf{\bar{x}}_{t} + \frac{\beta_t}{\alpha_t} \bigtriangledown_{\mathbf{\bar{x}}} \log{p_{t}(\mathbf{\bar{x}})} + \sqrt{\frac{\beta_t}{\alpha_t}} \ \epsilon_t \ \text{,}
\end{equation}
where $\log{p_{t}(\mathbf{\bar{x}})}$ is the score of $p_{t}$. From the \cite{song2020denoising}, we can get that $\bigtriangledown_{\mathbf{\bar{x}}} \log{p_{t}(\mathbf{\bar{x}})} = - \sqrt{\frac{1 - \alpha_{t}}{\alpha_{t}}} \ \epsilon_{\theta}^{(t)}(\alpha_{t} \mathbf{\bar{x}}_{t})$. Therefore, we can rewrite the Eq.~\ref{eq:DDIM} ($\eta = 1$) as:
{\small
\begin{alignat*}{4}
    \mathbf{\bar{x}}_{t-1} =& \ \mathbf{\bar{x}}_t + \sqrt{\frac{\alpha_{t}}{1 - \alpha_{t}}} \left( \frac{\alpha_{t} - \alpha_{t - 1}}{\alpha_{t - 1} \alpha_{t}} \right) \epsilon_{\theta}^{(t)}(\mathbf{x}_{t}) + \\
    & \sqrt{\frac{(1 - \alpha_{t - 1}) (\alpha_{t - 1} - \alpha_{t})}{(1 - \alpha_{t}) \alpha_{t - 1}^2}} \epsilon_t \\
    =& \ \mathbf{\bar{x}}_t + \frac{\beta_t}{\alpha_t} \bigtriangledown_{\mathbf{\bar{x}}} \log{p_{t}(\mathbf{\bar{x}})} + \sqrt{\frac{(1-\beta_t)(1-\beta_t-\alpha_t)\beta_t}{(1-\alpha_t)\alpha_t}} \ \epsilon_t \\
    \approx& \  \mathbf{\bar{x}}_t + \frac{\beta_t}{\alpha_t} \bigtriangledown_{\mathbf{\bar{x}}} \log{p_{t}(\mathbf{\bar{x}})} + \sqrt{\frac{\beta_t}{\alpha_t}} \ \epsilon_t
\end{alignat*}
}
Therefore, the sampler of Eq.~\ref{eq:DDIM} ($\eta = 0$, $1$ or $\sqrt{(1 - \alpha_{t})/(1 - \alpha_{t - 1})} $) used as the baseline in this paper is fundamentally another discretization to the same reverse-time SDEs/ODEs. We further propose the adaptive momentum sampler based on these DDPM/DDIM reverse process to improve the quality of generated images.

\subsection{Second-Order Approximation}

In this study, we present a mathematical demonstration that our proposed method, the Adaptive Momentum Sampler, converges to the true sample of the model. Furthermore, we conduct a preliminary mathematical analysis showing that the momentum sampler constitutes a second-order approximation of SDEs/ODEs. Consistent with Song et al. \cite{song2020score}, the DDPM/DDIM algorithm exhibits similarities to the Euler method for solving the following SDEs/ODEs:
\begin{equation}
    \mathbf{\bar{x}}_{t-1} = \mathbf{\bar{x}}_{t} + \ \mu \epsilon_{\theta}^{t}(\mathbf{x}_{t}) + \frac{\sigma_t}{\sqrt{\alpha_{t - 1}}} \cdot \epsilon_t  \ \mathrm{,}
    \label{eq:DDPM/DDIM}
\end{equation}
where {\small $\bar{\mathbf{x}}_t = \ \frac{\mathbf{x}_{t}}{\sqrt{\alpha_{t}}}$} and {\small $\mu = \left( \sqrt{\frac{1 - \alpha_{t - 1} - \sigma_t^2}{\alpha_{t - 1}}} - \sqrt{\frac{1 - \alpha_{t}}{\alpha_{t}}} \right)$}. We can rewrite Eq.~\ref{eq:DDPM/DDIM} as the first-order numerical procedure with $\mathbf{d \bar{x}}_t = \mathbf{\bar{x}}_{t} - \mathbf{\bar{x}}_{t-1}$:
\begin{equation}
    \mathbf{d \bar{x}}_t = -\mu \cdot \epsilon_{\theta}^{(t)}(\mathbf{x}_{t}) - \frac{\sigma_t}{\sqrt{\alpha_{t - 1}}} \cdot \epsilon_t \ \mathrm{,} 
    \label{eq:euler}
\end{equation}
Consider using a second-order numerical solver for DDPM/DDIM, the resulting equation is given by:
\begin{equation}
    \mathbf{d^2 \bar{x}}_t + \lambda \mathbf{d \bar{x}}_t = - \mu \cdot \epsilon_{\theta}^{(t)}(\mathbf{x}_{t}) - \frac{\sigma_t}{\sqrt{\alpha_{t - 1}}} \cdot \epsilon_t \ \mathrm{,} \label{eq:second-order}
\end{equation}
where the parameter $\lambda$ acts as a friction term. Furthermore, this former~\ref{eq:second-order} can be rewrite as: 
\begin{alignat*}{2}
    \mathbf{d \bar{x}}_t &= \eta \ \mathrm{,} \\
    -\mathbf{d^2 \bar{x}}_t &= -\mathbf{d \eta} = \lambda \eta + \mu \cdot \epsilon_{\theta}^{(t)}(\mathbf{x}_{t}) + \frac{\sigma_t}{\sqrt{\alpha_{t - 1}}} \cdot \epsilon_t \ \mathrm{.}
\end{alignat*}
With the midpoint method: 
{\small
\begin{alignat}{3}
    \mathbf{\bar{x}}_{t} - \mathbf{\bar{x}}_{t-1} =& \eta_{t - 0.5} \ \mathrm{,} \label{eq:midpoint_1} \\
    \eta_{t + 0.5} - \eta_{t - 0.5} =& \lambda \ \frac{\eta_{t + 0.5} + \ \eta_{t - 0.5}}{2} \nonumber \\
    &- \mu \epsilon_{\theta}^{t}(\mathbf{x}_{t}) -  \frac{\sigma_t}{\sqrt{\alpha_{t - 1}}} \cdot \epsilon_t \ \mathrm{,} \label{eq:midpoint_2}
\end{alignat}
}
we can reformulate these two Eq.~\ref{eq:midpoint_1} and  Eq.~\ref{eq:midpoint_2} to get the same Momentum Sampler formula as in main paper: 
\begin{alignat*}{2}
    \mathbf{\bar{x}}_{t-1} &= \enspace \mathbf{\bar{x}}_{t} + \mathbf{m}_{t-1} \ \mathrm{,} \\
    \mathbf{m}_{t-1} &= \enspace a \cdot \mathbf{m}_{t} + b \cdot (\mu \cdot \epsilon_{\theta}^{(t)}(\mathbf{x}_{t}) + \frac{\sigma_t}{\sqrt{\alpha_{t - 1}}} \cdot \epsilon_t) \ \mathrm{,}
\end{alignat*}
where $ \mathbf{m}_{t} = - \eta_{t + 0.5}$, $ a = (2 - \lambda) / (2 + \lambda)$ and $b = -2/(2+\lambda)$.
Moreover, by modifying the implicit difference method, it is clear that the our method is a second-order approximation of SEDs/ODEs.

\section{Samples}

\begin{figure}[!htpb]
    \vskip -0.1in
    \centering
    \begin{subfigure}[b]{0.6\linewidth}
    \centering
        \includegraphics[width=\linewidth]{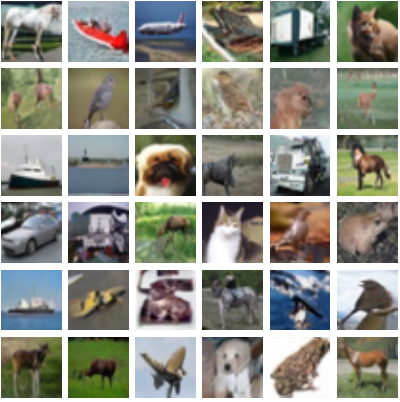}
        \vskip -0.1in
        \caption{DDIM-AM}
        \label{fig:DDIM-AM-cifar10}
    \end{subfigure}
    \begin{subfigure}[b]{0.6\linewidth}
     \centering
         \includegraphics[width=\linewidth]{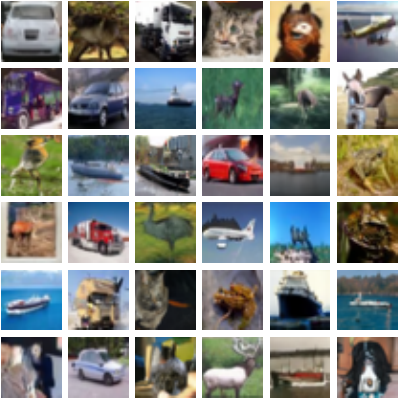}
         \vskip -0.1in
         \caption{DDPM-AM}
         \label{fig:DDPM-AM-cifar10}
    \end{subfigure}
    \vskip -0.1in
    \caption{CIFAR10 samples from 1000 step DDIM-AM (up) and DDPM-AM (down)}
    \label{fig:cifar10-AM}
    \vskip -0.2in
\end{figure}

\begin{figure*}[!htpb]
     \centering
     \begin{subfigure}[b]{\textwidth}
     \centering
        \includegraphics[width=0.95\textwidth]{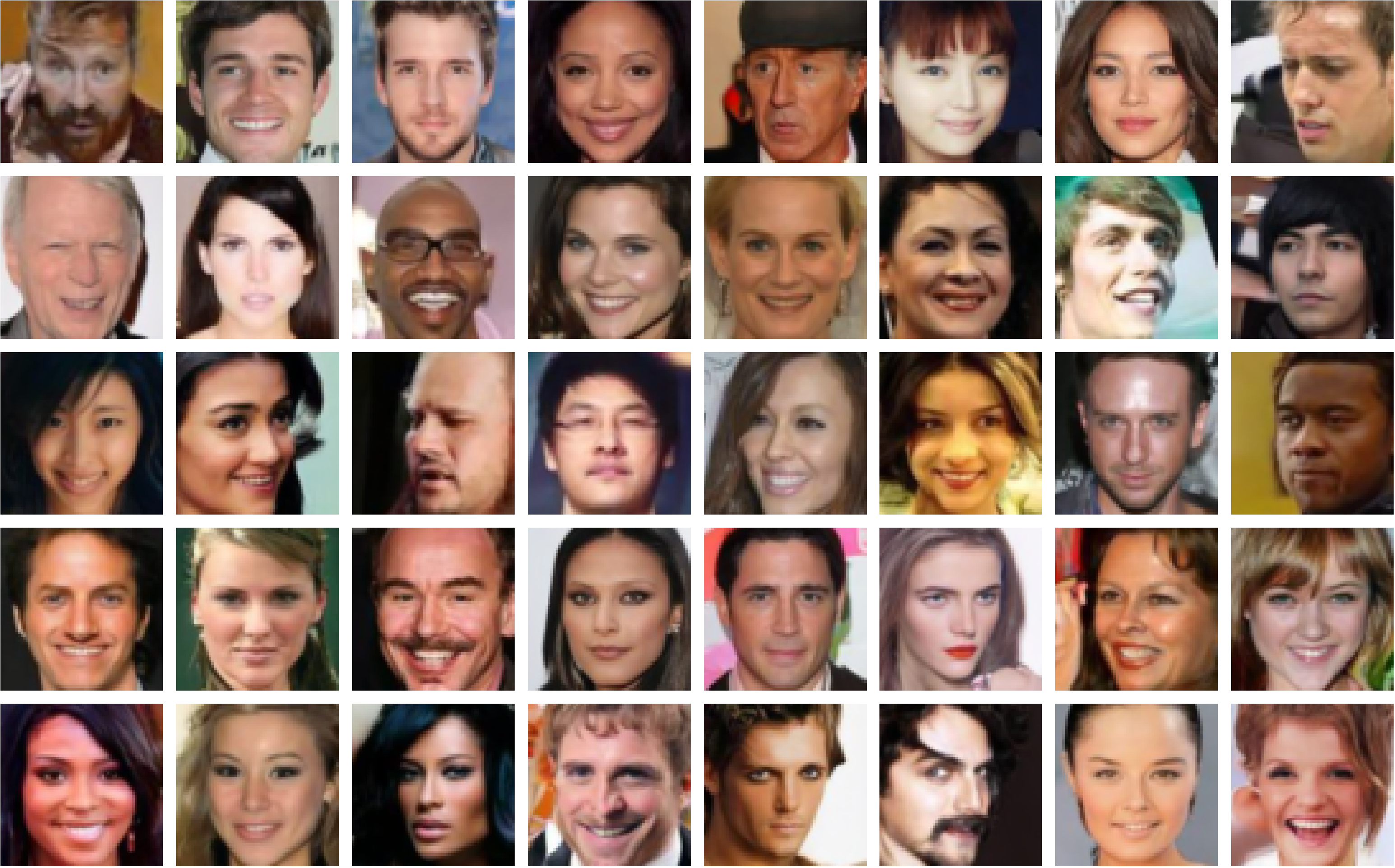}
        \caption{DDIM-AM}
        \label{fig:DDIM-AM-CelebA}
    \end{subfigure}
    \begin{subfigure}[b]{\textwidth}
     \centering
         \includegraphics[width=0.95\textwidth]{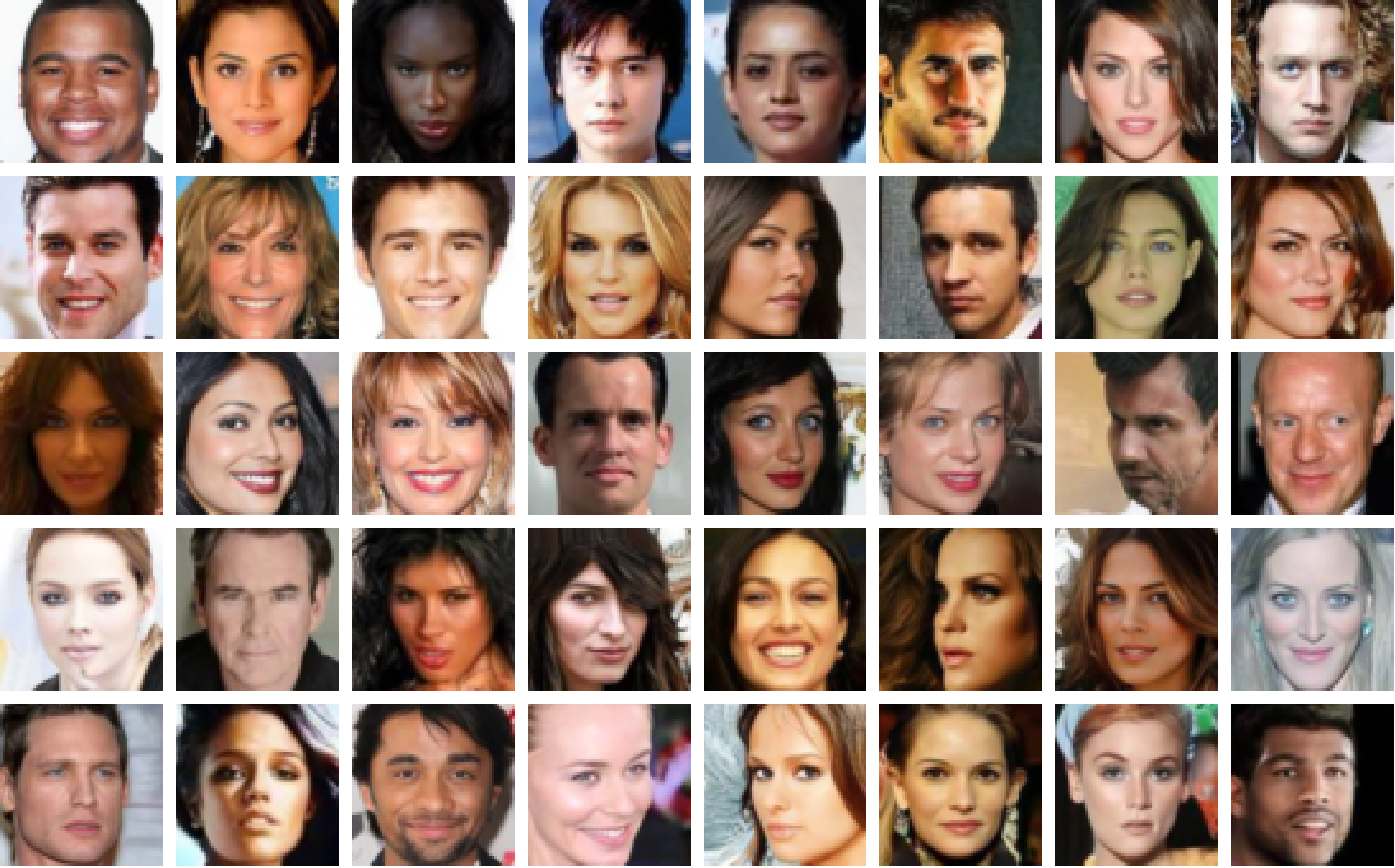}
         \caption{DDPM-AM}
         \label{fig:DDPM-AM-CelebA}
    \end{subfigure}
    \caption{CelebA samples from 1000 step DDIM-AM (up) and DDPM-AM (down)}
    \label{fig:CelebA-AM}
\end{figure*}

\begin{figure*}[!htpb]
     \centering
     \begin{subfigure}[b]{\textwidth}
     \centering
        \includegraphics[width=0.95\textwidth]{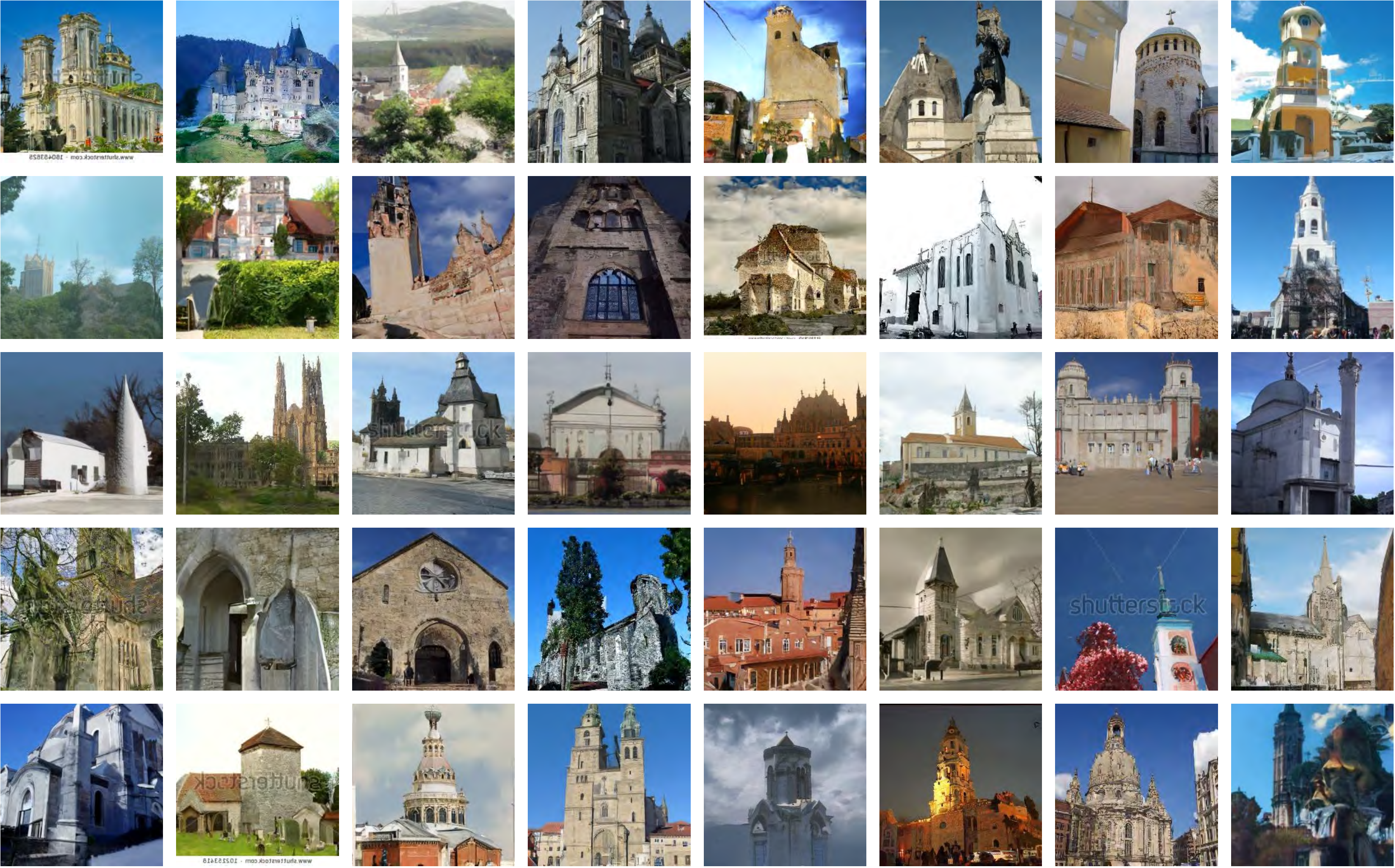}
        \caption{DDIM-AM}
        \label{fig:DDIM-AM-church}
    \end{subfigure}
    \begin{subfigure}[b]{\textwidth}
     \centering
         \includegraphics[width=0.95\textwidth]{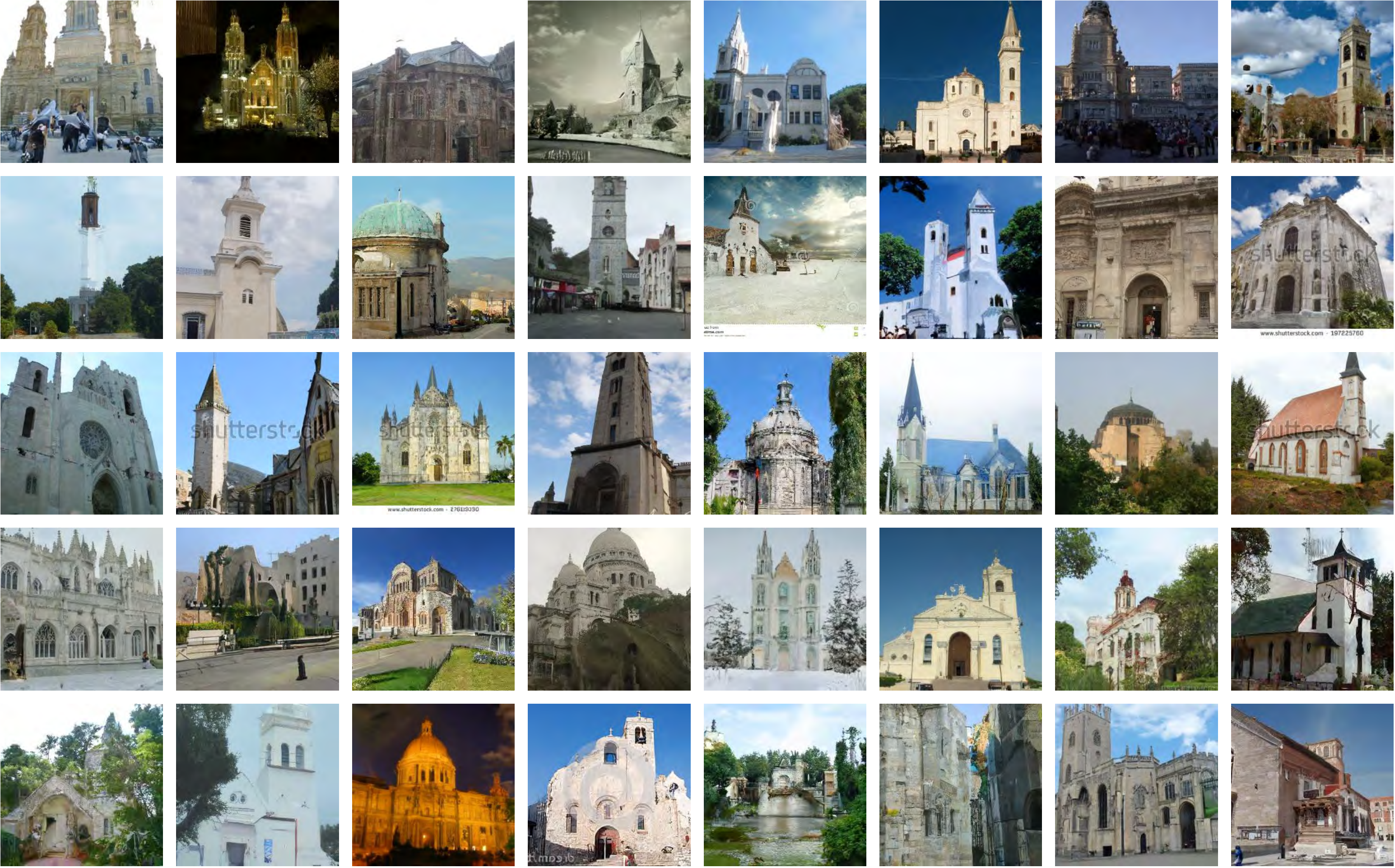}
         \caption{DDPM-AM}
         \label{fig:DDPM-AM-church}
    \end{subfigure}
    \caption{LSUN Church samples from 1000 step DDIM-AM (up) and DDPM-AM (down)}
    \label{fig:church-AM}
\end{figure*}

\begin{figure*}[!htpb]
     \centering
     \begin{subfigure}[b]{\textwidth}
     \centering
        \includegraphics[width=0.95\textwidth]{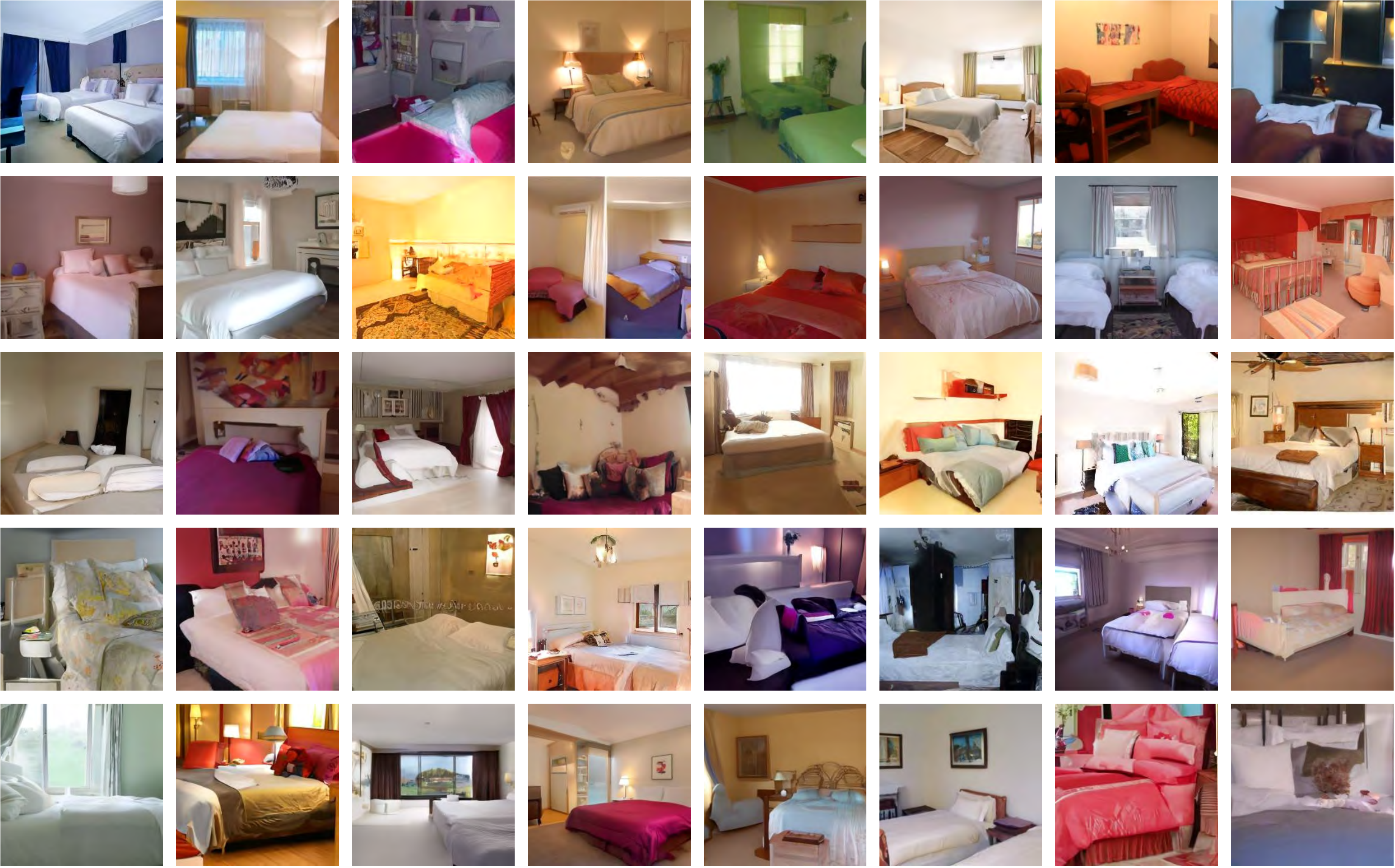}
        \caption{DDIM-AM}
        \label{fig:DDIM-AM-bedroom}
    \end{subfigure}
    \begin{subfigure}[b]{\textwidth}
     \centering
         \includegraphics[width=0.95\textwidth]{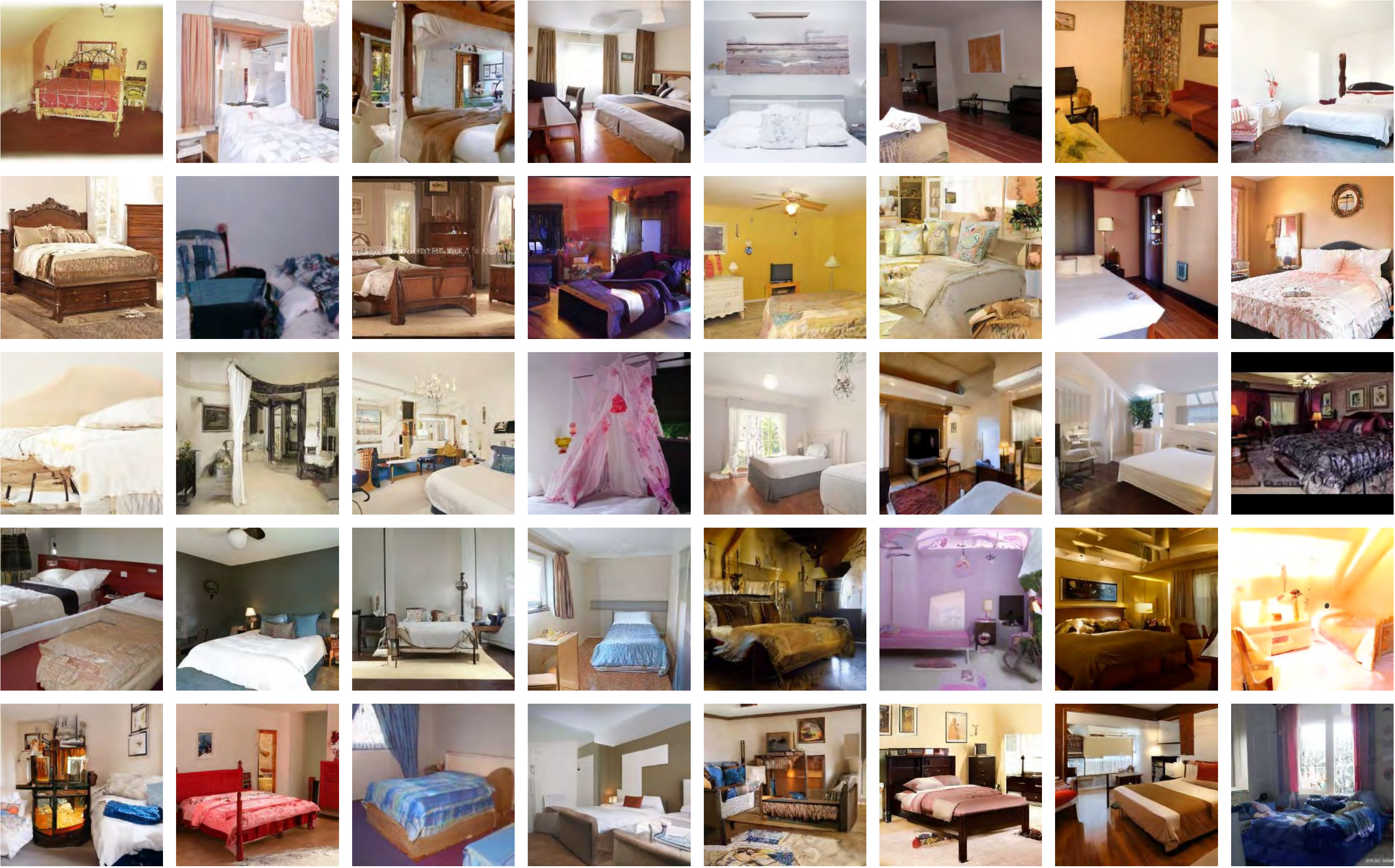}
         \caption{DDPM-AM}
         \label{fig:DDPM-AM-bedroom}
    \end{subfigure}
    \caption{LSUN Bedroom samples from 1000 step DDIM-AM (up) and DDPM-AM (down)}
    \label{fig:bedroom-AM}
\end{figure*}

In this section, we show more samples in Fig.~\ref{fig:cifar10-AM} (CIFAR10), Fig.~\ref{fig:CelebA-AM} (CelebA), Fig.~\ref{fig:church-AM} (Church) and Fig.~\ref{fig:bedroom-AM} (Bedroom).


\end{document}